\documentclass[lettersize,journal]{IEEEtran}
\usepackage{amsmath,amsfonts}
\usepackage{algorithmic}
\usepackage{array}
\usepackage[caption=false,font=normalsize,labelfont=sf,textfont=sf]{subfig}
\usepackage{textcomp}
\usepackage{stfloats}
\usepackage{lipsum}
\usepackage{url}
\usepackage{verbatim}
\usepackage{graphicx}
\usepackage{xtab}

\def\BibTeX{{\rm B\kern-.05em{\sc i\kern-.025em b}\kern-.08em
    T\kern-.1667em\lower.7ex\hbox{E}\kern-.125emX}}
\usepackage{balance}
\usepackage{graphicx} 
\begin{document}
\title{Random-coupled Neural Network}
\author{\emph{Haoran Liu, Mingzhe Liu, Peng Li, Jiahui Wu, Xin Jiang, Zhuo Zuo, Bingqi Liu}
\thanks{This work was supported by the National Natural Science Foundation of China under Grant U19A2086, Grant 12205078, and Grant 42104174. (Corresponding author: Mingzhe Liu).}
\thanks{Haoran Liu is with the State Key Laboratory of Geohazard Prevention and Geoenvironment Protection, Chengdu University of Technology, Chengdu 610059, China, and also with the Applied Nuclear Technology in Geosciences Key Laboratory of Sichuan Province, Chengdu University of Technology, Chengdu 610059, China (e-mail: liuhaoran@cdut.edu.cn).}
\thanks{Mingzhe Liu and Xin Jiang are with the State Key Laboratory of Geohazard Prevention and Geoenvironment Protection, Chengdu University of Technology, Chengdu 610059, China, and also with the School of Data Science and Artificial Intelligence, Wenzhou University of Technology, Wenzhou 325000, China (e-mail: liumz@cdut.edu.cn; jiangxin@cqut.edu.cn).}
\thanks{Peng Li is with the Engineering \& Technical College of Chengdu University of Technology, Leshan 614000, China (e-mail: lipeng@stu.cdut.edu.cn).}
\thanks{Jiahui Wu is with the Applied Nuclear Technology in Geosciences Key Laboratory of Sichuan Province, Chengdu University of Technology, Chengdu 610059, China (e-mail: wujiahui@stu.cdut.edu.cn).}
\thanks{Zhuo Zuo is with the Engineering \& Technical College of Chengdu University of Technology, Leshan 614000, China, and also with the Southwestern Institute of Physics, Chengdu 610225, China (e-mail: zuozhuo@stu.cdut.edu.cn).}
\thanks{Bingqi Liu is with the Chengdu University, Chengdu 610106, China (e-mail: liubingqi@cdu.edu.cn).}
\thanks{Color versions of one or more of the figures in this paper are available online at http://ieeexplore.ieee.org.}}

\maketitle

\begin{abstract}
\textbf{Improving the efficiency of current neural networks and modeling them in biological neural systems have become popular research directions in recent years. Pulse-coupled neural network (PCNN) is a well applicated model for imitating the computation characteristics of the human brain in computer vision and neural network fields. However, differences between the PCNN and biological neural systems remain: limited neural connection, high computational cost, and lack of stochastic property. In this study, random-coupled neural network (RCNN) is proposed. It overcomes these difficulties in PCNN’s neuromorphic computing via a random inactivation process. This process randomly closes some neural connections in the RCNN model, realized by the random inactivation weight matrix of link input. This releases the computational burden of PCNN, making it affordable to achieve vast neural connections. Furthermore, the image and video processing mechanisms of RCNN are researched. It encodes constant stimuli as periodic spike trains and periodic stimuli as chaotic spike trains, the same as biological neural information encoding characteristics. Finally, the RCNN is applicated to image segmentation, fusion, and pulse shape discrimination subtasks. It is demonstrated to be robust, efficient, and highly anti-noised, with outstanding performance in all applications mentioned above.
}
\end{abstract}


\section{Introduction}
\IEEEPARstart{S}{cientists} have been racing to study neuromorphic computing since the deep learning research boom caused by advances in graphics cards. Deep neural networks have achieved remarkable abilities today, for example, studying ecological networks \cite{RN1} and the famous AlphaGo that defeated a professional human Go player \cite{RN2}. However, these neural networks are computationally cumbersome and have high energy requirements compared to the biological neural system. Studies in neuroscience have found that the extraordinary abilities of the human brain come from three main aspects: a significant level of connectivity, time-dependent functionality, and structural and functional organizational hierarchy\cite{ RN3}. These neurological findings have shed light on the development of neuromorphic computing, leading to the research on third-generation neural networks.

Based on neuronal functionality, Maass divides neural networks into three generations \cite{RN4}. The first-generation is the perceptron proposed by McCulloch et al. \cite{RN5}, which constructs a linear relationship between the input and output signals. The second-generation includes all popular deep-learning models. A sigmoid unit or a rectified linear unit is added to the second-generation neural networks, introducing continuous nonlinearity to the neural networks. This nonlinearity extensively expanded the capable tasks of neural networks and essentially founded the modern neural networks discipline. The third-generation refers to the neural networks that process information via spikes, such as the spiking neural networks \cite{RN6, RN7, RN8} and pulse-coupled neural network \cite{RN9}. These third-generation neural networks imitate brains' information analysis strategy. They use the neurons as the computational primitive elements, memorize and learn through the neurons (specifically, the synapses integrated with neurons), and transfer information via discrete spikes. The pulse-coupled neural network (PCNN), as a representative third-generation model, has been applied in numerous fields, such as color image processing \cite{RN10, RN11}, automatic diagnosis \cite{RN12, RN13}, image fusion \cite{RN14}, computer vision \cite{RN15}, pulse shape discrimination \cite{RN16}, and image enhancement \cite{RN17}.

Although researchers have made many efforts to improve the biomimicry of the PCNN \cite{RN18}, several significant differences between the PCNN and the actual visual cortex remain. A few differences include (1) extensive connectivity between neurons in the brain and limited neighboring connections in PCNN; (2) the brain’s high computational efficiency and PCNN’s high computational burden; and (3) inherently stochastic spike activities in the brain and deterministic calculations in PCNN. In the current PCNN models, a central neuron is only connected to eight neighboring neurons, a minimal number of connections compared with the human brain. The link condition is curtailed by the computational ability of computers. In the human brain, spike transmission is calculated through the highly efficient analog circuit; continuous integration and decay of action potential only happen within the neurons while information is exchanged through discrete spike transmission. In contrast, the computation in the PCNN model is through digital circuits, which require a large amount of computation. Although the PCNN utilize the same efficient continuous-discrete separated computation manner, the expensive digital computing leads to severe computing cost. Furthermore, neurology research has demonstrated that the stochastic characteristics of the human brain play a vital role in the information processing task\cite{RN19}. However, the PCNN is based on deterministic Boolean circuits. As long as the input and parameters are given, the neuron activity of PCNN is deterministic.

In 2022, Liu et al. proposed the continuous-coupled neural network (CCNN), which adds the stochastic property to PCNN by modifying the neuronal excitation process \cite{RN20}. They demonstrated that CCNN could achieve better image and video processing abilities with a stochastic neuronal excitation process. However, the spikes generated by the CCNN are no longer discrete signals, which deviates from the brain-like computation and leads to additional computational burden.

In this study, a novel PCNN architecture is introduced. It is the random-coupled neural network (RCNN). The RCNN realizes stochastic properties while reserving the discrete characteristics of spike transmissions, which is inherently different from the CCNN. In the RCNN model, neuron connectivity is significantly expanded; a central neuron is connected to more than 20 neighboring neurons. A stochastic process called random inactivation compensates for the computational burden caused by this vast connection. The inactivation closes some neural connections randomly, making a central neuron capable of obtaining information from many neurons while remaining at an affordable computation level. The RCNN model solves three significant drawbacks in the PCNN models simultaneously: limited neural connection, high computational cost, and lack of stochastic property.

It is found that RCNN encodes constant stimuli as periodic spike trains and periodic stimuli as chaotic spike trains, the same information encoding mechanism as the biological neural systems. On this basis, we conduct experiments in image segmentation, image fusion, and pulse shape discrimination fields, researching the two-dimensional and one-dimensional signal processing performance of RCNN. Additionally, RCNN’s image and video processing characteristics are researched. 

The layout of this study is arranged as follows: related works are presented in Section 2; RCNN’s architecture and image and video processing mechanism are illustrated in Section 3; in Section 4, experiments were conducted in three different applications, and experimental results are analyzed; in Section 5, the conclusion of this study is drawn, and future research is discussed.

Contributions of this study:
\begin{list}{}{}
	
	\item 1) Drawbacks of PCNN that prevent it from achieving high standard neuromorphic computing are analyzed;
	\item 2) Radom-coupled neural network is proposed, which considers the link between neurons as a stochastic process, solving PCNN’s three drawbacks simultaneously;
	\item 3) RCNN’s information encoding mechanism is researched;
	\item 4) The proposed RCNN is applied in image segmentation, fusion, and pulse shape discrimination, resulting in robust and outstanding performance. Its vast potential to be used in numerous fields is demonstrated.
\end{list}

\section{Related Works}
\subsection{Pulse-coupled neural network}
In 1994, Johnson et al. proposed PCNN under the inspiration of the biological neural cortex \cite{RN9}. They further developed the initial artificial cortical model \cite{RN21}and applied it to the image processing field, opening a new era of brain-like computing and digital image processing. PCNN processes information via spikes, inherited from the spike transmission process between cell assemblies. Like the human brain, PCNN uses individual neurons as the information processing unit. Neurons compute and store information through the charge potential’s growth and decay, while thresholds control the communication between neurons. If the charge potential of a neuron exceeds its threshold, this neuron is activated and will pass spikes to its connected neurons. It is worth noticing that the threshold of each neuron is not fixed. All thresholds are under dynamic fluctuations, making neurons exhibit different properties under different stimuli. The PCNN is based on temporal coding, which extracts and presents an image's information by the neurons' ignition counts. This temporal coding mechanism has been demonstrated to be very important for fast neural information processing \cite{RN22,RN23}.

The computation between neurons is discrete; either 1 (spike generated) or 0 (no spike generated) are transmitted. The continuous mathematical process only happens within each neuron, leading to impressive computational efficiency compared to second-generation neural networks. PCNN receives multiple stimuli from an image when this image is fed as an input to the PCNN. After many iterations, the neurons record the count of their ignition. A neuron’s ignition count contains the extracted information from this neuron’s corresponding location in the input image. The mathematical expressions of PCNN are given as follows:
\begin{equation}
	F_{ij}[n] = V_{F} \sum_{kl} M_{ijkl} Y_{kl}[n-1] + S_{ij}\\
	+e^{-\alpha_{F}}F_{ij}[n-1]
\end{equation}
\begin{equation}
    L_{ij}[n] = e^{-\alpha_{L}} L_{ij}[n-1]+V_{L}\sum_{kl}W_{ijkl}Y_{kl}[n-1]
\end{equation}
\begin{equation}
	U_{ij}[n] = F_{ij}[n] (1 + \beta L_{ij}[n])
\end{equation}
\begin{equation}
	\theta_{ij} [n] = e^{-\alpha_{\theta}}\theta_{ij}[n-1]+V_{\theta}Y_{ij}[n-1]
\end{equation}
\begin{equation}
	Y_{ij}[n] = \left\{
	\begin{array}{l}
		1, U_{ij}[n] > \theta_{ij}[n] \\
		0, \text{ otherwise}
	\end{array}
	\right.
\end{equation}
where, 
$ F_{ij} $ and $ L_{ij} $ are the feedback input and link input of a neuron located at $ (i,j) $ , respectively; 
$ U $ denotes a neuron’s internal activity (i.e., membrane potential); 
$ \theta $ is the dynamic threshold;
$ Y $ is the output of a neuron (i.e., the spike generation condition); 
$ n $ denotes the iteration count; 
$ \alpha_{F} $ and $ \alpha_{L} $ are constant decay coefficients of the feedback input and link input, while $ V_{F} $and $ V_{L} $ are constant weighting coefficients of the feedback input and link input; $ M_{ijkl} $ and $ W_{ijkl} $ are weight matrixes that control the connection between a central neuron located at $ (i,j) $ and its neighboring neurons located at $ (k,l) $; S denotes the external input fed to PCNN; $\beta $ is a weighting factor that controls the relationship between feedback and link inputs; $ \alpha_{\theta} $ and $ V_{\theta} $ are constant decay coefficient and constant weighting coefficient of the dynamic threshold, respectively.

According to the mathematical formula of PCNN, the feedback input $ F $, link input$ L $, internal activity$ U $, and dynamic threshold$ \theta $are inseparably connected. The relationship and activity between and within them are modulated by several constant decay coefficients ($ \alpha_{F} $, $ \alpha_{L} $, and  $ \alpha_{\theta} $) and weighting coefficients ($ V_{F} $,  $ V_{L} $, and $ V_{\theta} $). When the PCNN is used to process an external stimulus S, the four parts above of a neuron fluctuate under several iterations. During this dynamic fluctuation, this neuron produces spikes and passes them to its connected neurons if the spike generation condition is satisfied ($ U_{ij} [n]>$ $\theta_{ij} [n] $). After a given iteration count, the ignition counts of neurons corresponding to different parts of the input image exhibit different patterns. Because the number of neurons in PCNN is the same as the number of pixels in the input image and there is a one-to-one correspondence between neurons and pixels, all these ignition counts compose an ignition map with the same size as the input image. This ignition map contains extracted information that can be further used in applications such as image segmentation and fusion.

\subsection{Simplified pulse-coupled neural network}
The simplified pulse-coupled neural network (SPCNN) \cite{RN24} is a further evolved model based on Zhan et al.’s spiking cortical model \cite{RN25}. It significantly reduces the number of manual parameters compared to the PCNN and has high accuracy. The intrinsic information processing manner of SPCNN is the same as that of PCNN mentioned in the last section. As one of the most popular PCNN-derived models in image processing field, it has been applied to numerous applications such as object recognition \cite{RN26} and multi-disciplinary image segmentation \cite{RN24,RN25}. The formula of SPCNN are given as follows:
\begin{equation}
	U_{ij}[n] = S_{ij}(1+\beta V_{U} \sum_{kl} {M}_{ijkl} Y_{kl}[n-1])+e^{-\alpha_{U}}\\ U_{ij}[n-1] 
\end{equation}
\begin{equation}
	Y_{ij}[n] = \left\{
	\begin{array}{l}
		1, U_{ij}[n] > \theta_{ij}[n] \\
		0, \text{ otherwise}
	\end{array}
	\right.
\end{equation}
\begin{equation}
	\theta_{ij}[n] = e^{-\alpha_{\theta}}\theta_{ij}[n-1]+V_{\theta}Y_{ij}[n-1]
\end{equation}
where all the notations have the same meaning as indicated in PCNN’s mathematical expressions; the differences between the PCNN and SPCNN are:  (1) Feedback input and link input are simplified and integrated to the internal activity $U$ and (2) Constant decay coefficient $\alpha_U$ and constant weighting coefficient $V_U$ are added to modulate characteristics of internal activity $U$.

\subsection{Image segmentation}

Image segmentation uses a computer algorithm to delineate anatomical structures, highlighted objects, and other regions of interest. It has been used in numerous fields, such as computer-aided diagnosis \cite{RN29} , object recognition \cite{RN30} , pattern recognition \cite{RN31,RN32,RN33}, digital photography \cite{RN35} and cell image segmentation \cite{RN36,RN37,RN38}.
\subsubsection{Otsu}

In 1979, Otsu proposed an automatic threshold selection method [39]. This method is simple and robust; hence has been widely used in various image segmentation applications. The Otsu method determines a threshold by maximizing the weighted sum of between-class variances of foreground and background pixels. Specifically, the threshold determination process is as follows:
	\begin{enumerate}
		\item The gray value of an image is divided into two categories:
		\begin{equation}
			\left\{
			\begin{array}{l}
				G_1 = \{0,1,2,\ldots,T\} \\
				G_2 = \{T+1,T+2,\ldots,L-1\}
			\end{array}
			\right.
		\end{equation}
		where, $ G1 $ denotes the group of gray level that are smaller than $ T $; $ G2 $ denotes the group of gray level that are bigger than $ T $; L is the image’s total number of the gray levels.
		\item The probabilities of gray level appearing in $ G1 $ and $ G2 $ are calculated via the formula as follows:
		\begin{equation}
			P_{G1} = \sum_{i=0}^T p_{i}
		\end{equation}
		\begin{equation}
			P_{G2} = 1-P_{G1}
		\end{equation}
		where, $ p_{i} $denotes the probability of occurrence of gray level$ i $, defined as the number of pixels at gray level i divided by the total number of pixels.
		\item The means of $ G1 $ and $ G2 $ are calculated as:
		\begin{equation}
			\mu_{G1} = \sum_{i=0}^T \frac{i*p_{i}}{P_{G1}}
		\end{equation}
		\begin{equation}
			\mu_{G2} = \sum_{i=T+1}^{L-1 }\frac{i*p_{i}}{P_{G2}}
		\end{equation}
		\item Finally, the optimal threshold T  is determined by maximizing the between-class variance:
		\begin{equation}
		\tilde{T}=\operatorname{Arg} \max _{0<T<L-1} \sigma^{2}(T)
		\end{equation}
		where,
		\begin{equation}
			\sigma^2(T) = P_{G1}P_{G2}(\mu_{G1}-\mu_{G2})^2
		\end{equation}
		The Otsu is an efficient and robust image segmentation method in most applications. However, this method is not resistant to noise and has specific requirements for the image's gray-level distribution. It performs poorly when an image contains too much noise, or an image's histogram has multiple peaks.
	\end{enumerate}
\subsubsection{K-mean}
The K-mean is a traditional statistical image segmentation method\cite{RN40,RN41}. It separates the image’s grey-level intensities into several clusters, categorizing pixels into different clusters based on the difference between its grey-level intensity and the cluster’s central grey-level intensity. This categorization process is performed in an iteration manner, which means that the clusters and the central grey-level intensity of a cluster change through iteration. The segmentation process is done when all clusters tend to be stable, and no more pixels fluctuate from cluster to cluster. Specifically, the K-mean segmentation process is as follows:
	\begin{enumerate}
		\item  K clusters are arbitrarily selected, with cluster centers defined;
		\item  The grey-level intensity difference between each pixel and cluster center is calculated. Categorize each pixel into the cluster with minimum grey-level intensity difference;
		\item Recalculate cluster centers;
		\item Iterate through steps (2) to (3) until all pixels are stable in a single cluster.
	\end{enumerate}
	
	The K-mean method can process segmentation tasks that traditional methods with a fixed threshold cannot handle. However, it is entirely based on image grey-level intensity, leading to poor performance when several different objects with the same grey-level intensity exist.
\subsubsection{PCNN}

The PCNN and its derived models have been widely used in the image segmentation field due to their outstanding information extraction capability\cite{RN24,RN38,RN42}. Unlike deep learning neural networks, PCNN has a single layer and processes information without pre-training. This single layer of the PCNN is a two-dimensional matrix of linked neurons with the same dimension as the input image. Each neuron in this matrix corresponds to a pixel in the input image. PCNN extracts information from the input image through an iterative manner, which feeds the input image to the networks several times and encodes information into the ignition state of neurons (their ignition counts after multiple iterations). The output of PCNN, the ignition map, includes ignition counts of all neurons and can be further used to image processing. 

In image segmentation, the ignition state of the last iteration (neurons corresponding to pixels in the input image can be either fired or not fired) is used for segmentation, or the total ignition counts of neurons are used for segmentation with a threshold operation.
\subsection{Image fusion}
The image fusion is the process of extracting information from multiple images and presenting all information in a single one \cite{RN43,RN44}. These information fusion algorithms have been developed in numerous fields, such as infrared and visible light image fusion\cite{RN45}, multi-focus image fusion \cite{RN46}, nuclear magnetic resonance imaging and computed tomography image fusion \cite{RN47}, and multichannel image fusion of x-ray differential phase contrast imaging \cite{RN48}.
\subsubsection{Multiscale image decomposition}
The multiscale image decomposition technique utilizes as much information as possible in an image, realizing high-quality image processing. Early multiscale decomposition approaches used a bilateral filter to preserve edges in the image-smoothing task \cite{RN49,RN50}. More recent multiscale decomposition algorithms are based on domain transforms, such as Fourier transform \cite{RN51}, wavelet transform \cite{RN52}, nonsubsampled contourlet transform \cite{RN53,RN54}, and nonsubsampled shearlet transform \cite{RN55,RN56}.

These domain transform algorithms decompose images into different scales or even further into different directions. Commonly, an image can be decomposed into several sub-band images with the same size as the original one. Fine details, textures, edges, and structures of an image are easier to access in decomposed images.

In image fusion, two or three input images are decomposed into several sub-band images, and a fusion rule is used to select pixels from these sub-bands. The selected pixels contain more information than unselected ones at the corresponding location in the input images. All selected pixels compose a group of sub-band images and can be further reconstructed into the final fused image.
\subsubsection{PCNN}
The PCNN is commonly used as a pixel selection tool in the image fusion application \cite{RN57,RN59}. The information of corresponding pixels in different input images or images’ decomposition sub-bands is evaluated by PCNN. The PCNN processes these images or sub-bands with the same parameter settings, outputting several ignition maps. The ignition count of a pixel’s corresponding neuron indicates the importance of its information. Pixels with higher ignition counts are selected and fused into a final image or decomposition sub-bands under a given fusion rule.

As a result of PCNN’s information extraction ability and the biological cortex’s information processing style, it considers the general importance of a pixel’s information, including details, textures, and edges. This information processing capability leads to a better image fusion performance than those methods with pixel select strategy of direct pixel value comparison.
\subsection{Pulse shape discrimination}
The pulse shape discrimination algorithm aims to separate neutron and gamma-ray pulse signals \cite{RN60,RN61}. These pulse signals are one-dimensional arrays and highly similar, leading to difficulties in neutron detection because the gamma-ray photons inevitably accompany neutrons. Traditional discrimination methods incorporate, for example, charge comparison \cite{RN62} and zero crossing \cite{RN63,RN64}, extracting pulse signals’ information through integration or integral-differential process. These methods find the difference between the falling edge and delayed fluorescence parts of pulse signals where neutron and gamma-ray pulses differ most \cite{RN65}.

After the information extraction process, a discrimination factor is calculated for each radiation pulse signal. It is used to draw a histogram (the X-axis is the discrimination factor, and the Y-axis is the pulse signal count), in which the counts of pulse signals distribute as two separate groups. One group incorporates gamma-ray pulse signals, and the other includes neutron pulse signals.
\subsubsection{PCNN}
In 2021, Liu et al. used the PCNN to discriminate neutron and gamma-ray pulse signals \cite{RN66}. They demonstrated that the PCNN could efficiently extract the pulse signal's information, generating an ignition map for each pulse signal. This information extraction process of PCNN considers dynamic information inside pulse signals and has outstanding anti-noise capability \cite{RN16,RN67}. It is similar to the PCNN's general consideration of information's importance in image fusion applications. In pulse shape discrimination, PCNN considers not only the signal's amplitude but also the signal's changing pattern and decaying rate. Intuitively, the difference between the ignition maps of pulse signals is much more significant than that of the original signal. After extracting information from PCNN or PCNN-derived models, a discrimination factor calculation rule (for example, integration of the ignition map or ladder gradient calculation) is applied to ignition maps. When the discrimination factors are calculated, further discrimination processes are the same as the traditional discrimination methods.
\begin{figure}[t]
	\centering
	\includegraphics{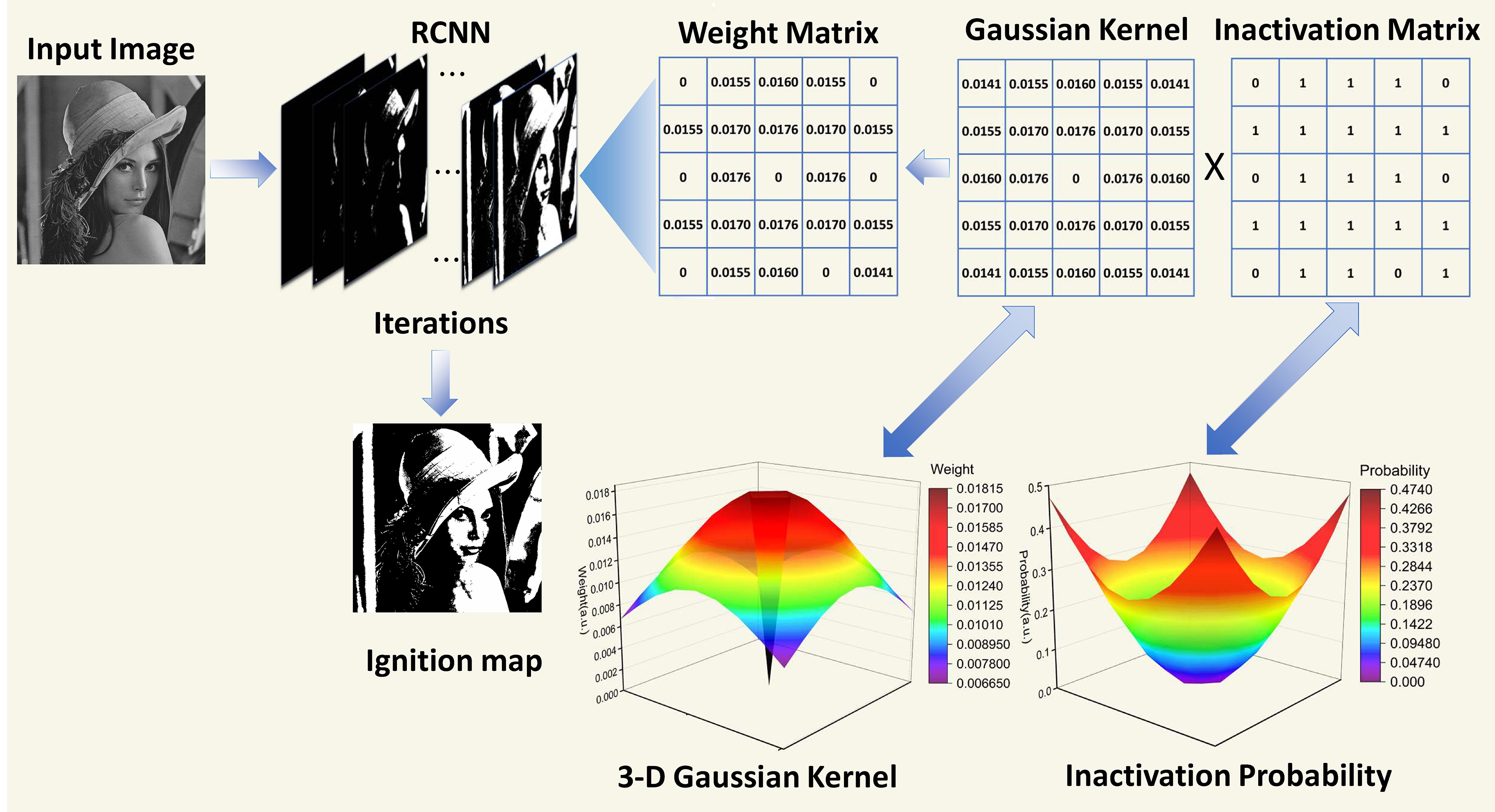}
	\caption{Signal processing scheme of the random-coupled neural network. The input image is used as an external stimulus that iteratively stimulates the RCNN multiple times. The weight matrix of RCNN is composed of Hadamard products between the Gaussian kernel and inactivation matrix, which both follow Gaussian distributions. This weight matrix is constructed so that neurons further away from the central neuron have lower weights and higher inactivation probabilities, and vice versa. After multiple iterations of RCNN processing, the ignition counts of each pixel are recorded and presented in the ignition map. Important information such as details, edges, and textures are transported from the original input image to the ignition map, which can be further used for various image processing applications.}
\end{figure}
\section{Methodology}
\subsection{Problems in current PCNN-derived models}
The neural response characteristic is a significant distinction between the standard Hodgkin-Huxley neuronal model and most PCNN-derived models. The biological neural response is stochastic, while that of most PCNN-derived models is deterministic. In 2022, Liu et al. added randomness to the PCNN model to achieve a stochastic neural response. They considered pulse generation as a stochastic process that follows a Gaussian distribution. This novel network structure is called the continuous-coupled neural network (CCNN).

CCNN exhibits periodic characteristics under constant stimuli, while it exhibits chaotic characteristics under periodic stimuli. This neural response is consistent with biological neural properties. However, the CCNN’s pulse generator becomes continuous under the consideration of randomness. This phenomenon leads to a major difference between the natural neural structure and the CCNN architecture. The real neural cortex processes information continuously inside neurons while transforming information discretely between neurons. In contrast, CCNN transforms information continuously between neurons.

In other words, CCNN achieves stochastic properties at the expense of brain-like computations' discrete inter-neuron information transfer properties and computational efficiency.  Only one problem out of three of PCNN’s major drawbacks is solved, which are limited neural connection, high computational cost, and lack of stochastic property.
\subsection{Modeling of random-coupled neural network}
The stochastic response of a neural structure is undoubtedly vital for brain-like computing, while the discrete information exchange between neurons is also an essential characteristic of neuromorphic computing. Consequently, in the random-coupled neural network (RCNN), we consider adding randomness into the PCNN architecture in a different aspect compared to the CCNN. On the one hand, the discrete pulse generator is reserved, making neurons communicate via discrete spikes (0 or 1). On the other hand, the stochastic neural response is achieved by a random inactivation weight matrix of link input. The signal processing scheme of the random-coupled neural network is shown in Fig. 1.

As shown in Fig. 1, the RCNN processes information through multiple iterations and encodes information via spike trains. The total ignition counts of neurons consist of the ignition map, which contains extracted features and can be further used to image processing applications. This information process is the same as the original PCNN model. The ability to communicate discretely between neurons is preserved. However, the weight matrix that controls the link between a central neuron and its surrounding neurons is designed to be stochastic. It is constructed by the Hadamard product (elemental-wise multiplication) of a gaussian kernel and an inactivation matrix. The gaussian kernel determines the link strength between neurons. It follows a two dimensional Gaussian distribution: $ N~(0,0,4^2,4^2,0) $. The mathematical definition of a two-dimensional Gaussian distribution is given as follows,
\begin{equation}
	\begin{split}
		f(x,y) = & \frac{1}{2\pi\sigma_{1}\sigma_{2}\sqrt{1-\rho^2}} \exp \left[ -\frac{1}{2(1-\rho^2)} \right. \\
		& \left. \left( \frac{(x-\mu_{1})^2}{\sigma_{1}^2} - \frac{2\rho(x-\mu_{1})(y-\mu_{2})}{\sigma_{1}\sigma_{2}} + \frac{(y-\mu_{2})^2}{\sigma_{2}^2} \right) \right]
	\end{split}
\end{equation}
where,  
$ f(x,y) $ is the probability of a point in a two-dimension space; 
$ \mu_{1} $ and $ \mu_{2} $ are means; 
$ \sigma_{1} $ and $ \sigma_{2} $ are variances; 
and $ \rho $ is the correlation. A two-dimensional Gaussian distribution can be uniquely defined as $ N~(\mu_{1},\mu_{2},\sigma_1^2,\sigma_2^2, \rho) $. The central point of the gaussian kernel is set to 0 as the central neuron cannot receive link stimuli from itself. The greater the distance between neurons and the central neuron, the weaker the stimulus to the central neuron. The inactivation matrix is composed of 0 and 1, and a position's value is 1 when the weight channel is open and 0 when the weight channel is closed. 

To construct the inactivation matrix mentioned above, the probability of a channel closing, called the inactivation probability, must first be defined. The inactivation probabilities of all channels compose the drop out probability matrix, which has the same size as the inactivation matrix, and it follows a Gaussian distribution: $ N~(0,0,5^2,5^2,0) $. The three-dimensional diagram of it is shown in Fig. 1. The greater the distance between neurons and the central neuron, the more likely the channel to close. Then, this inactivation probability matrix is normalized so that the elements in it are all in the range from 0 to 1. Finally, for each iteration of the RCNN, a matrix with the exact size of the inactivation matrix is generated randomly, in which all elements are random numbers in the range from 0 to 1, following a uniform distribution. As the size of the inactivation matrix is the same as that of the inactivation probability matrix, corresponding elements in these two matrixes can be compared one by one. If an element from the inactivation probability is bigger than its corresponding one from the generated random matrix, the corresponding weight channel in the inactivation matrix is closed and verse versa.
\begin{figure}[t]
	\centering
	\includegraphics{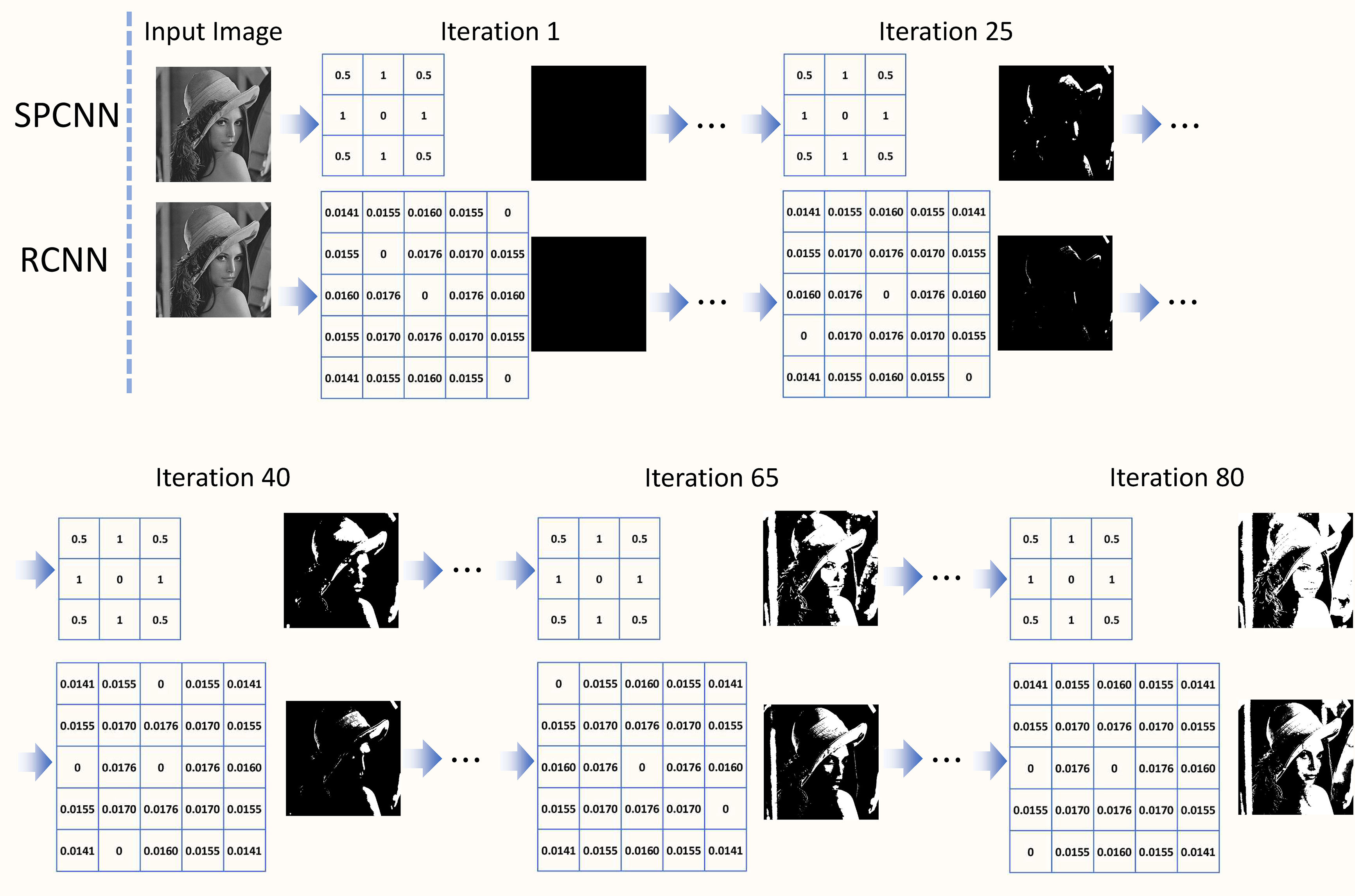}
	\caption{Weigh matrix characteristics of SPCNN and RCNN. The weight matrix of SPCNN is a fixed 3×3 matrix, linking the central neuron to eight surrounding neurons. However, the weight matrix of RCNN has a much larger size (5×5 here for illustration purposes, it can be expanded to 20×20 while still computationally affordable) and is constantly changing throughout iterations. Randomly closed weights release the computational burden of RCNN, making it capable of more vast neuron connectivity. Meanwhile, this inactivation process introduces stochastic properties to the RCNN, empowering it with better accuracy, robustness, and biologically plausible neural response.}
\end{figure}
\begin{figure}[t]
	\centering
	\includegraphics[width=0.5\textwidth]{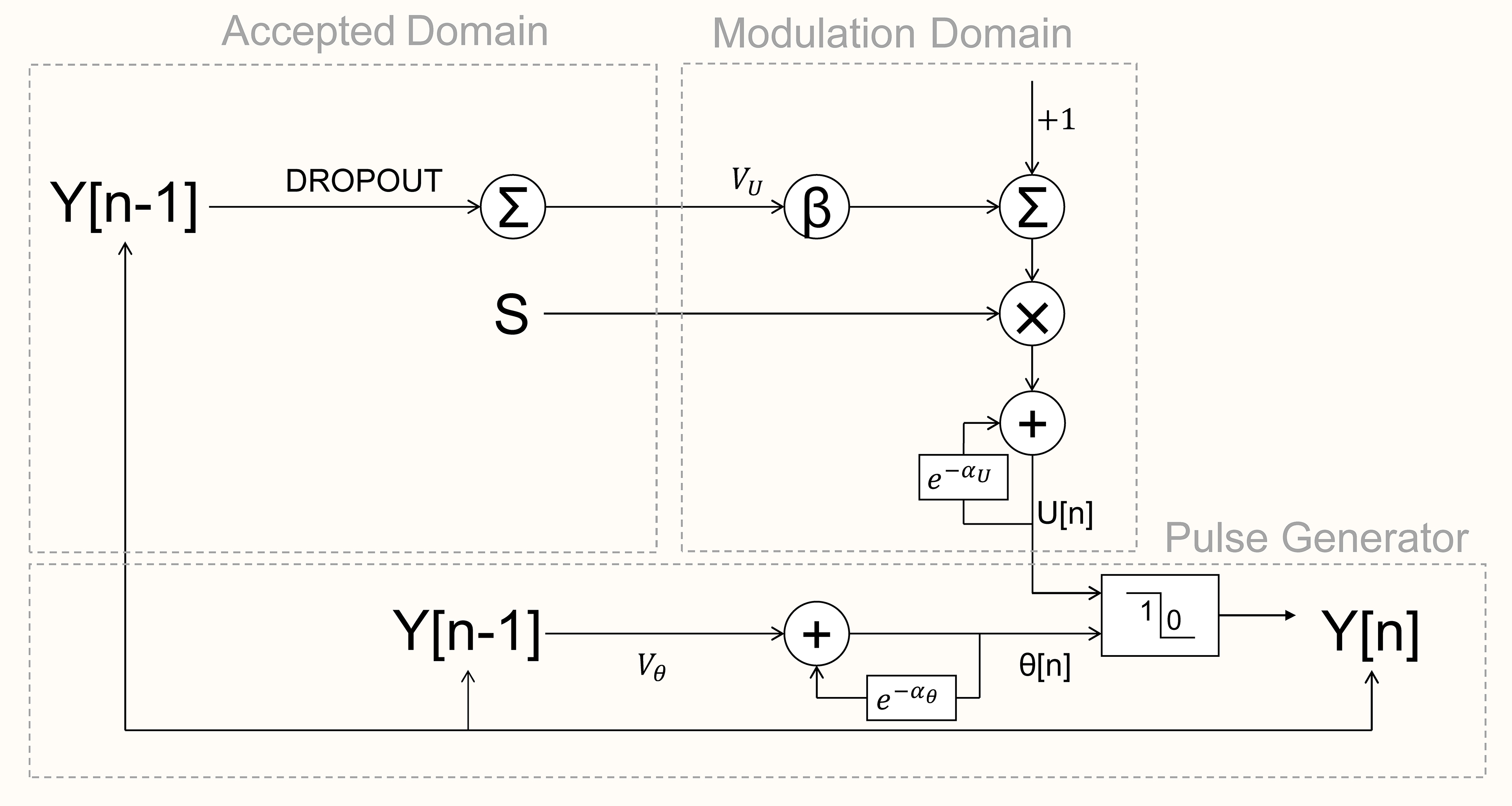}
	\caption{RCNN structure. The accepted domain receives external and link input stimuli. These stimuli contribute to the membrane potential, which is calculated in the modulation domain. The membrane potential and dynamic threshold decide the spike generation process in the pulse generator. RCNN has only two kinds of output: spike or no spike.}
\end{figure}
To further illustrate the weight matrix properties of the RCNN, the matrix difference between the SPCNN and RCNN is shown in Fig. 2. The weight matrix of SPCNN is a fixed 3×3 matrix, linking the central neuron to eight surrounding neurons. This neuron link is very narrow compared to the number of biological neuron dendrites. The vast connectivity of the brain’s neurons has been demonstrated to be vital in information processing. This limited connection of SPCNN is caused by the tradeoff between computational burden and neuromorphic computing accuracy. In contrast, the weight matrix of RCNN has a size of 5×5 (only for illustration purposes, it can be a matrix with any size), linking the central neuron to more than 20 surrounding neurons. The connection strength remains unchanged throughout the iterations, while some weight matrix channels are randomly closed in each iteration. This random inactivation makes the link input of RCNN exhibits stochastic characteristic and releases the computational burden of a vast link input. Released computational complicity and computer advances make a vast link input weight matrix possible. The size of RCNN’s weight matrix can be from 5×5 to 20×20. In conclusion, RCNN solves PCNN models’ problems of limited neural connection, high computational cost, and lack of stochastic property all at once.

The mathematical expressions of RCNN are given as follows,
\begin{equation}
	U_{ij}[n] = S_{ij}(1+\beta V_{U}\sum_{kl}\dddot{M}_{ijkl}Y_{kl}[n-1])+e^{-{\alpha_U}}U_{ij}[n-1]
\end{equation}
\begin{equation}
	Y_{ij}[n] = \left\{
	\begin{array}{l}
		1, U_{ij}[n] > \theta_{ij}[n] \\
		0, \text{ otherwise}
	\end{array}
	\right.
\end{equation}
\begin{equation}
	\theta_{ij}[n] = e^{-\alpha_{\theta}} \theta_{ij}[n-1]+V_{\theta}Y_{ij}[n-1]
\end{equation}
\begin{equation}
	\dddot{M}_{ijkl} = G_{ijkl} \cdot D_{ijkl}
\end{equation}
where, $\dddot{M}_{ijkl}$ is the random inactivation weight matrix, calculated by the Hadamard product (elemental-wise multiplication) between $G_{ijkl}$ and $D_{ijkl}$;$G_{ijkl}$is the Gaussian kernel, which follows a Gaussian distribution; $D_{ijkl}$ is the inactivation matrix, the inactivation probability of which also follows a Gaussian distribution; other notations have the same meaning as indicated in SPCNN’s mathematical expressions.

The network structure of RCNN is shown in Fig. 3. RCNN includes three major composites: accepted domain, modulation domain, and pulse generator. The accepted domain receives stimuli from neighboring neurons (i.e., the link input modulated by the inactivation process) and external stimulus S. These two kinds of stimulus are coupled in the modulation domain, constituting the membrane potential U. Finally, the membrane potential and dynamic threshold are compared in the pulse generator, in which spikes are produced if the pulse generation condition described in Equation (17) is satisfied. All information is contained in the precise spike time, while the amplitude of the spike represents no information.
\subsection{Characteristics in image and video processing}
In 2022, Liu et al. demonstrated the importance of CCNN’s different responses to constant and periodic stimuli \cite{RN20}. CCNN exhibits periodic characteristics under constant stimuli, with periods inversely proportional to stimulus intensity. In contrast, CCNN generates chaotic spike trains under periotic stimuli, which may achieve moving target recognition by analyzing the periodicity of output spike trains directly. Unlike current feature extraction-based object recognition methods, CCNN’s video processing ability is direct, robust, and highly anti-noise. In this section, the image and video processing properties of RCNN are researched. It is demonstrated that RCNN also encodes constant stimuli as periodic spike trains and periodic stimuli as chaotic spike trains. This information encoding characteristic is the same as that of CCNN and biological neuron networks. Parameter settings of RCNN for the image and video processing experiments are given in the Supplementary material A.
\begin{figure}[t]
	\centering
	\includegraphics[width=0.5\textwidth]{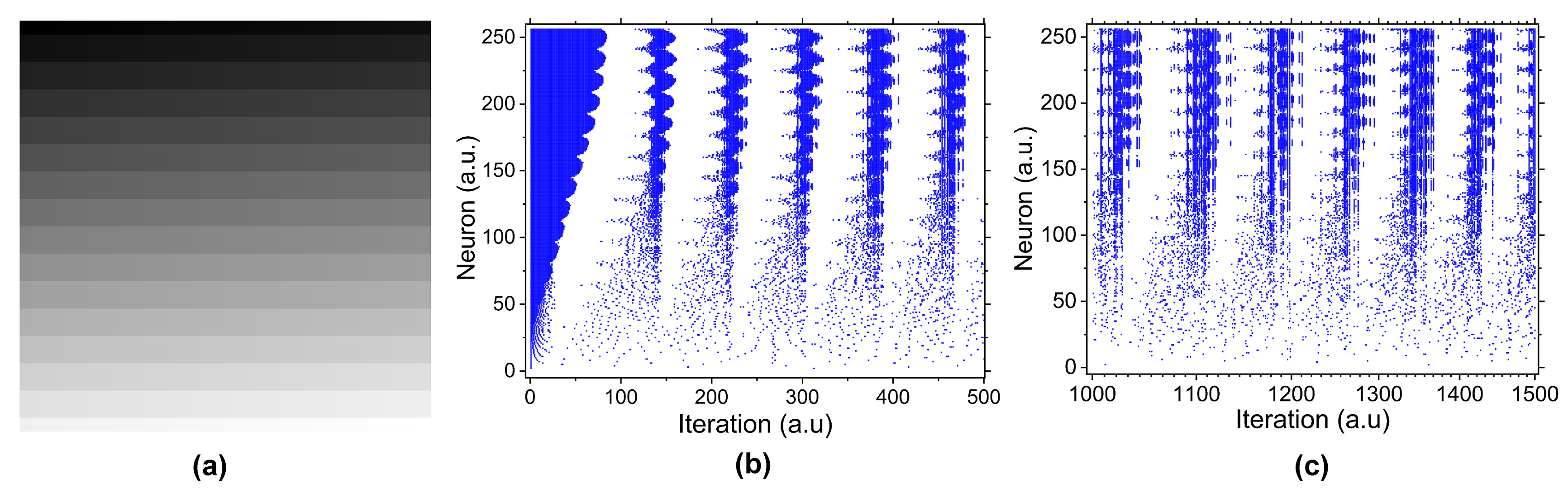}
	\caption{Image processing characteristics of RCNN. (a) Input stimuli, a 16×16matrix with a gray value of 0 at the upper left corner, a gray value of 255 at the lower right corner, and gradually changed gray values in the center. (b) Spike trains of the first 500 iterations. Y-axis denotes neurons that received different gray values, and a blue dot denotes a spike is generated. (c) Spike trains from 1000 to 1500 iterations. Y-axis denotes neurons that received different gray values, and a blue dot denotes a spike is generated.}
\end{figure}
\begin{figure}[t]
	\centering
	\includegraphics[width=0.5\textwidth]{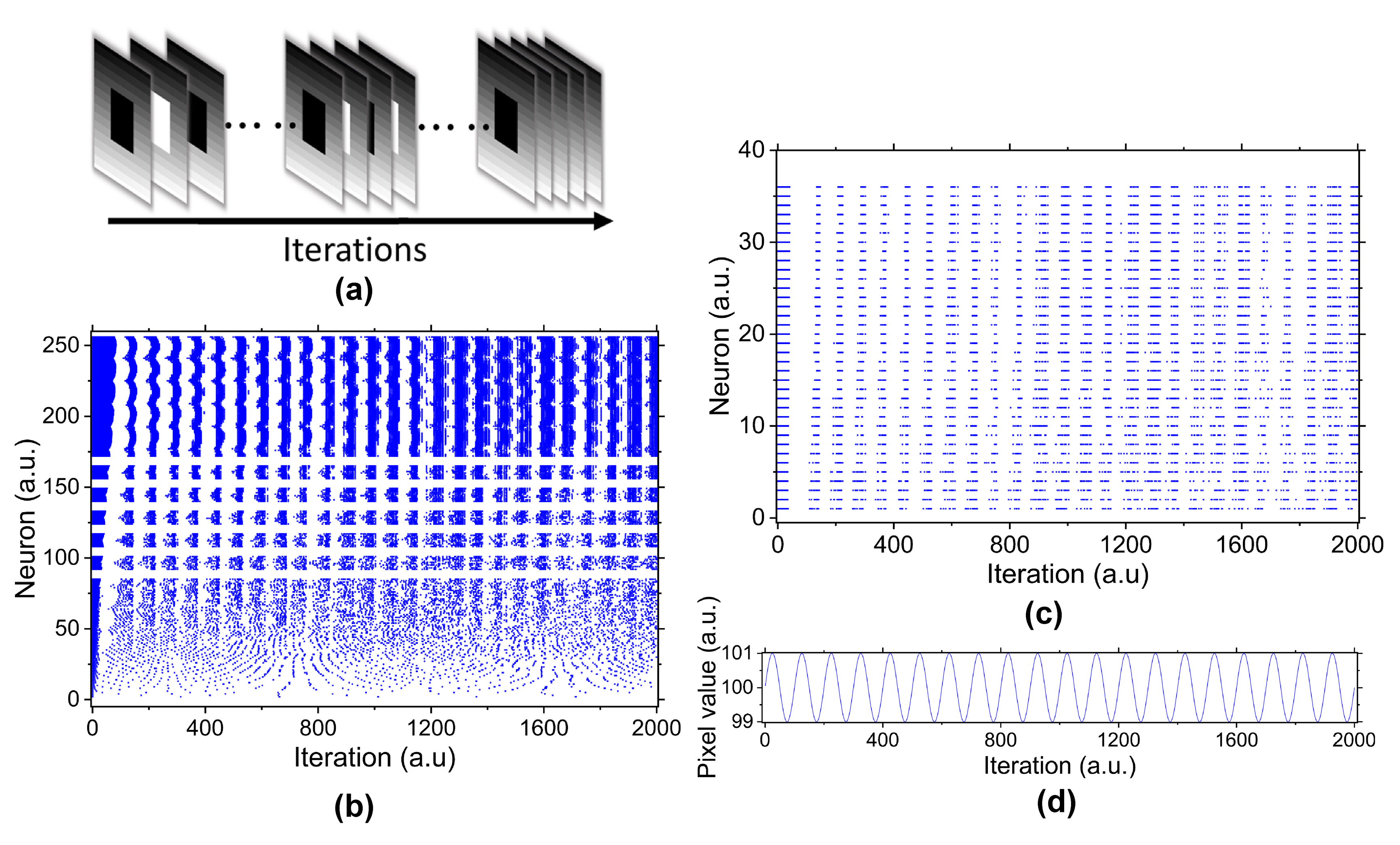}
	\caption{Video processing characteristics of RCNN. (a) Input stimuli, a 16×16 matrix with a 6×6 dynamic changing matrix in the center and the rest parts are the same as Fig. 4a. Stimuli x of dynamic changing matrix follow $ x=100+sin(n\pi⁄50)$, where n is the iteration count. Dynamic changing pixels are denoted by the alternating black and white area. (b) Spike trains of neurons that received constant stimuli. Y-axis denotes neurons that received different gray values, and a blue dot denotes a spike is generated. (c) Spike trains of neurons that received periodic stimuli. Y-axis denotes neurons, and a blue dot denotes a spike is generated. (d) Stimuli changing pattern of a neuron in the dynamic changing matrix.}
\end{figure}
\subsubsection{Image processing}
The experimental results of RCNN’s spike generation under constant stimuli are shown in Fig. 4. The neurons that received low gray values return to a resting state faster than those that received high gray values, as shown in Fig. 4b. After multiple iterations, RCNN’s spike trains reach stable and exhibits clear periodic characteristics, as shown in Fig. 4c. Moreover, the larger the gray value a neuron received, the smaller the neuron’s spike train period.
\subsubsection{Video processing}
The experimental results of RCNN’s spike generation in video processing are shown in Fig. 5. The video input is an image sequence with dynamic changing pixel values in the center and surrounding fixed pixel values. Fig. 5b shows the spike trains of neurons that received constant stimuli. Periodic properties that are the same as image processing scenarios can be seen. Fig. 5c is the neuron activities under periodic stimuli that$ follows x=100+sin(n\pi⁄50) $. There are periodic fluctuations in spike trains that are consistent with the changing pattern of stimuli shown in Fig. 5d. However, chaotic spike trains can be observed within each fluctuation period. This phenomenon demonstrates that RCNN encodes fixed pixels as periodic spike trains while encoding changing pixels as chaotic spike trains.

In conclusion, the characteristics of RCNN’s neuron activities are the same as that of the RCNN and biological neural networks. It realizes these characteristics through a stochastic process called random inactivation. The random inactivation weight matrix of link input adds stochastic properties to the RCNN architecture. It also releases the computational burden of link input, hence making the linking between a large number of surrounding neurons possible. Additionally, RCNN also achieves high accuracy and anti-noise ability, which will be illustrated in the next section.
\section{Applications}
In this section, RCNN’s signal processing performance is researched in three sub-tasks: image segmentation, image fusion, and pulse shape discrimination. The first two applications are currently popular research hot zone of digital image processing. The original PCNN and its derived models have been extensively used in these two fields. The efficiency and robustness of RCNN in these two fields are of great interest to the readers. Additionally, the pulse shape discrimination field is a newly emerged application direction for PCNN and its derived models. The advances of RCNN in this field will shed light on PCNN’s one-dimensional signal processing research.
\subsection{Image segmentation }
Four images were used in image segmentation experiments, some of which are selected from image datasets \cite{RN68} and \cite{RN69}. In image segmentation, the ignition counts of RCNN are categorized into two groups using the entropy maximizing method \cite{RN70}. These two groups are set as black and white pixels, and further as the final segmentation results. The performance of the RCNN-based segmentation method is compared to two traditional image segmentation methods, a SPCNN-based method, and a CCNN-based method \cite{RN20}. It is worth noting that the SPCNN-based method proposed by Chen et al. is capable for hierarchical segmentation with automatic parameter settings \cite{RN24}. Here we only segment images into two layers: the foreground layer and the background layer. Consequently, the SPCNN-based method can only be iterated twice, generating the segmentation results of two layers. Further hierarchical segmentation with multiple layers is not applied. Detailed parameters of all methods are given in the Supplementary material B, and the segmentation results are shown in Fig. 6-9.
\begin{figure}[t]
	\centering
	\includegraphics[width=0.5\textwidth]{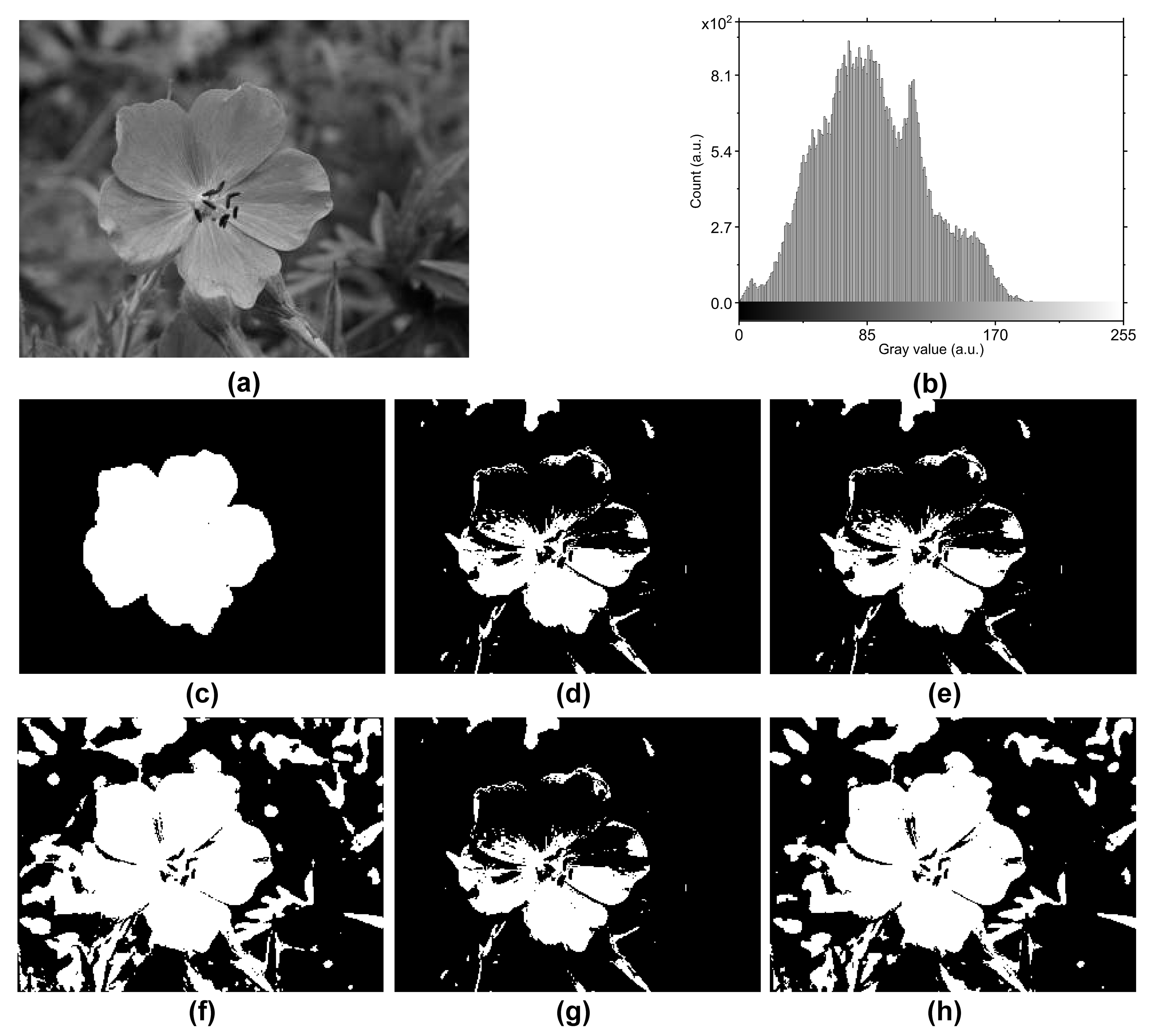}
	\caption{ Segmentation results of flower. (a) Original image. (b) Histogram of original image. (c) Ground truth. (d) Otsu. (e) K-mean. (f) SPCNN. (g) CCNN. (h) RCNN.}
\end{figure}
\begin{figure}[t]
	\centering
	\includegraphics[width=0.5\textwidth]{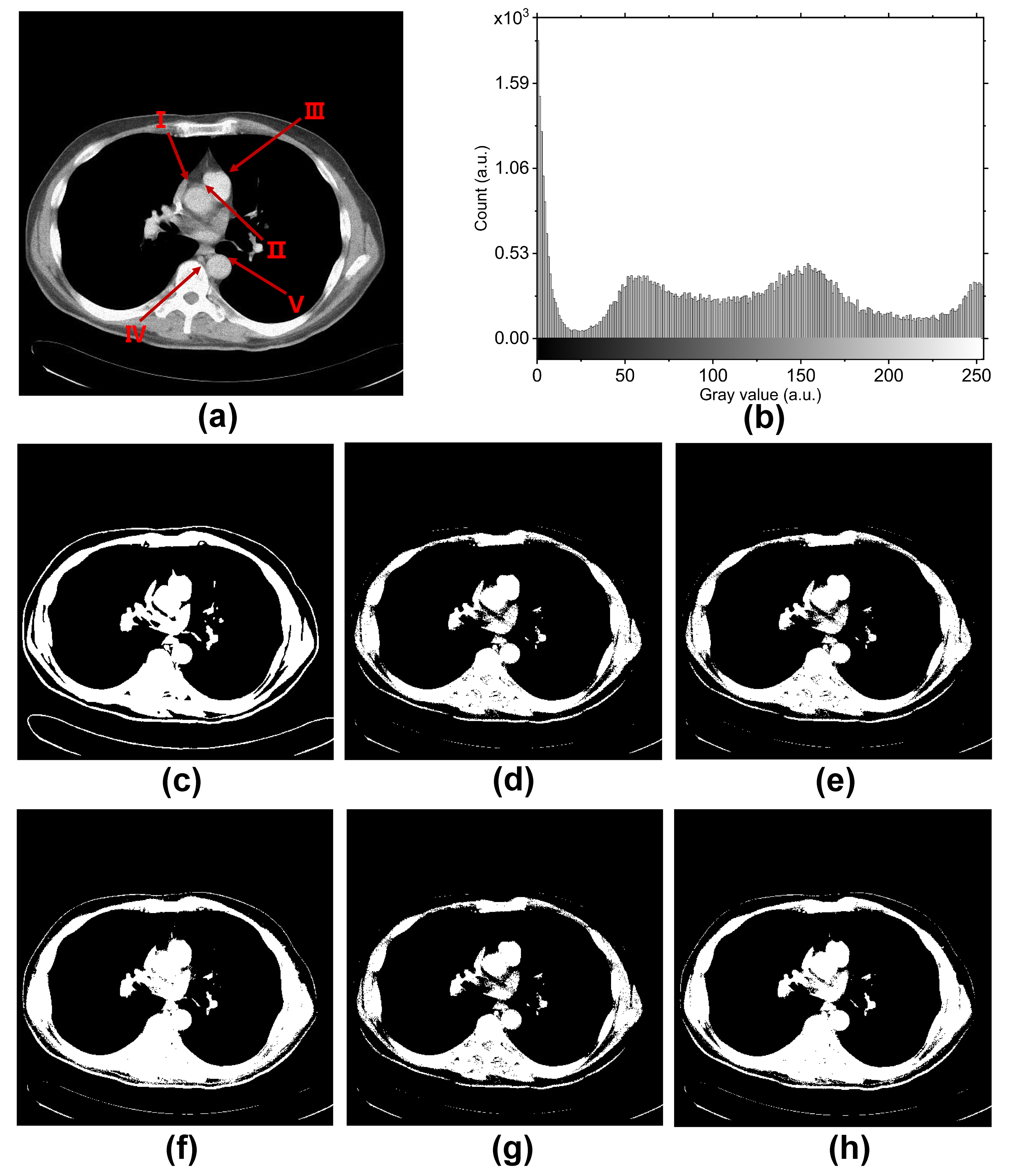}
	\caption{Segmentation results of CT image. (a) Original image. Axial section of chest CT with right atrium (I), ascending aorta (II), pulmonary artery (III), thoracic vertebra (IV), and descending aorta (V). (b) Histogram of original image. (c) Ground truth. (d) Otsu. (e) K-mean. (f) SPCNN. (g) CCNN. (h) RCNN.}
\end{figure}

Fig. 6a depicts an image with a histogram exhibiting a single peak. Most image segmentation algorithms perform effectively under this histogram condition. The ground truth of  the segmentation is illustrated in Fig. 6c, showcasing a flower as the foreground layer and the surrounding area as the background. Due to the lack of consistency in grayscale inside the flower and background, traditional segmentation approaches displayed in Fig. 6d and 6e fail to accurately represent the complete front layer. These methods solely rely on the distribution of grayscale values, disregarding important image features such as texture and edges. Additionally, the performance of CCNN is subpar, as evidenced in Fig. 6f, attributable to its continuous output nature. The ignition map produced by CCNN closely resembles the original image, leading to outcomes akin to traditional methods when thresholding is applied to it. In contrast, SPCNN and RCNN present a more cohesive front layer in Fig. 6f and 6h, despite significant variations in grayscale values within this layer. These methods recognize the front layer as a unified entity based on the features of the flower. While the background layer in SPCNN and RCNN is still impacted by notable fluctuations in grayscale values. Only machine learning algorithms can excel in addressing such semantic classification tasks. Notably, the incorporation of image features by SPCNN and RCNN highlights their potential to serve as effective feature extractors, enhancing machine learning algorithms in semantic classification tasks.

Fig. 7a presents a CT image with additional Poisson noise introduced using the MATLAB built-in function imnoise. In X-ray imaging, noise following the Poisson distribution is common, especially in scenarios with low photon flux. Fig. 7d-g reveal the unsatisfactory results of separating the right atrium (I) from the ascending aorta (II), also with difficulties in distinguishing between the ascending aorta (II) and pulmonary artery (III), and the thoracic vertebra (IV) and descending aorta (V). However, in Fig. 7h, the RCNN-based method excels in separating these anatomical structures, displaying a clear boundary between the thoracic vertebra (IV) and descending aorta (V). Overall, RCNN demonstrates superior performance compared to alternative methods in this context.

Fig. 8a and 9a represent extreme scenarios for binary image segmentation, where all pixel values are heavily clustered into distinct groups, leaving minimal pixels at other gray values, as depicted in histograms (Fig. 8b and 9b). Traditional segmentation techniques struggle to process such images, as evident in Fig. 8d, 8e, 9d, and 9e. Although hierarchical methods can typically segment images with such histogram characteristics into multiple layers, our focus here is on evaluating binary segmentation performance. Unfortunately, both SPCNN and CCNN yield unsatisfactory outcomes (Fig. 8f, 8g, 9f, and 9g), mirroring the limitations of traditional methods. These models fail to extract and analyze sufficient image features necessary for accurately identifying the foreground objects, such as airplanes and apples, with their segmentation decisions heavily influenced by variations in image pixels’ gray scale values. Conversely, the RCNN-based approach effectively isolates objects from the background (Fig. 8h and 9h), successfully mitigating the negative impact of wide grayscale ranges representing different pixel groups in the histogram. This robust performance is attributed to the extensive connectivity inherent in RCNN, where the collaboration of multiple linked neurons compensates for background disparities, preventing misclassification of dark grey backgrounds as foreground objects.

Objective evaluation of the aforementioned segmentation tasks is presented in Table 1. Evaluation criteria includes pixel accuracy (PA), intersection over union (IOU), and Dice coefficient \cite{RN71}.

In conclusion, the RCNN-based image segmentation method exhibits outstanding performance in all experiments mentioned above. It presents segmented images with little noise and clear boundaries. Moreover, it can segment images with special histograms, outperforming the others both subjectively and objectively.

\begin{figure}[t]
	\centering
	\includegraphics[width=0.5\textwidth]{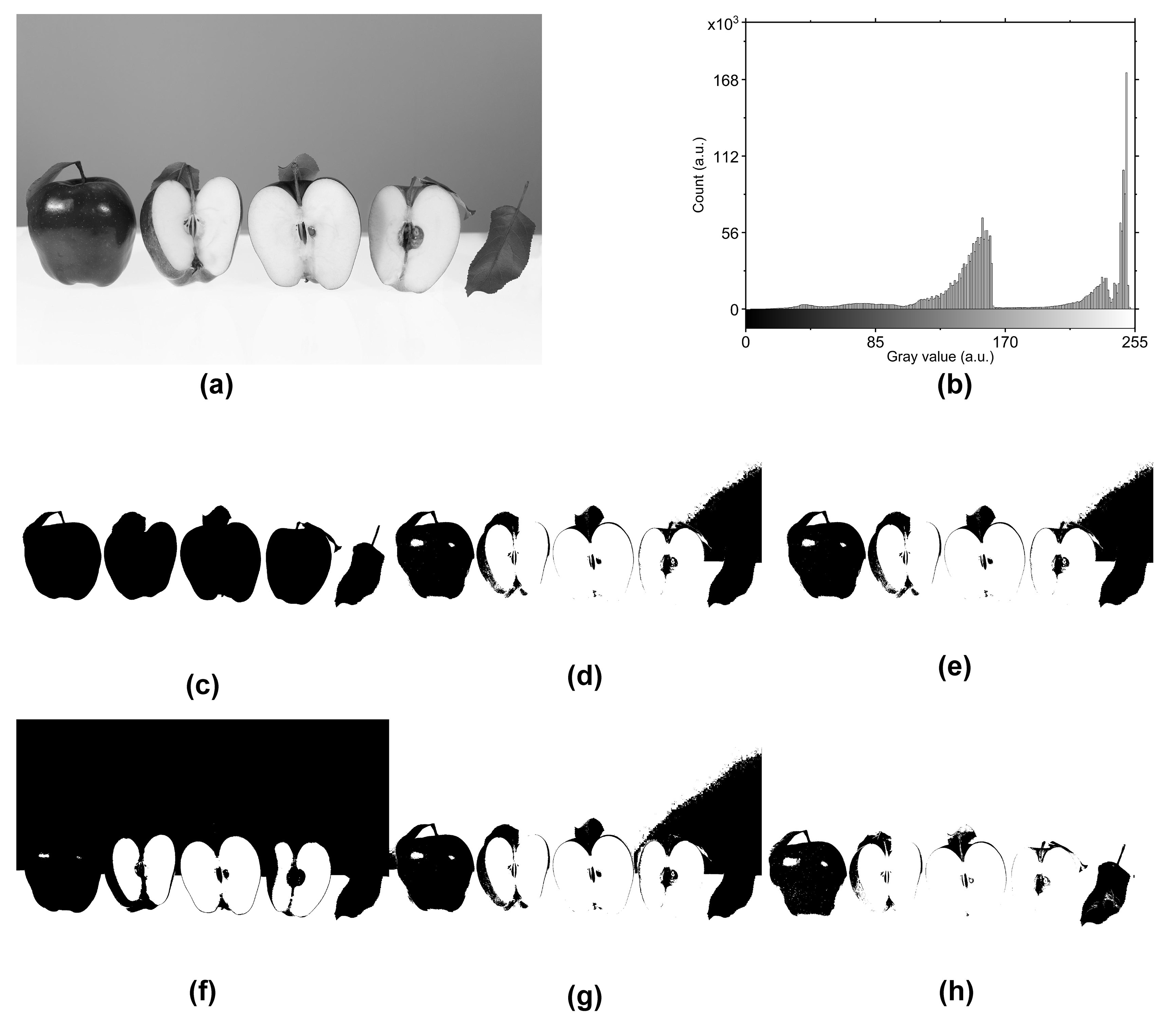}
	\caption{Segmentation results of apples. (a) Original image. (b) Histogram of original image. (c) Ground truth. (d) Otsu. (e) K-mean. (f) SPCNN. (g) CCNN. (h) RCNN.}
\end{figure}
\begin{figure}[t]
	\centering
	\includegraphics[width=0.5\textwidth]{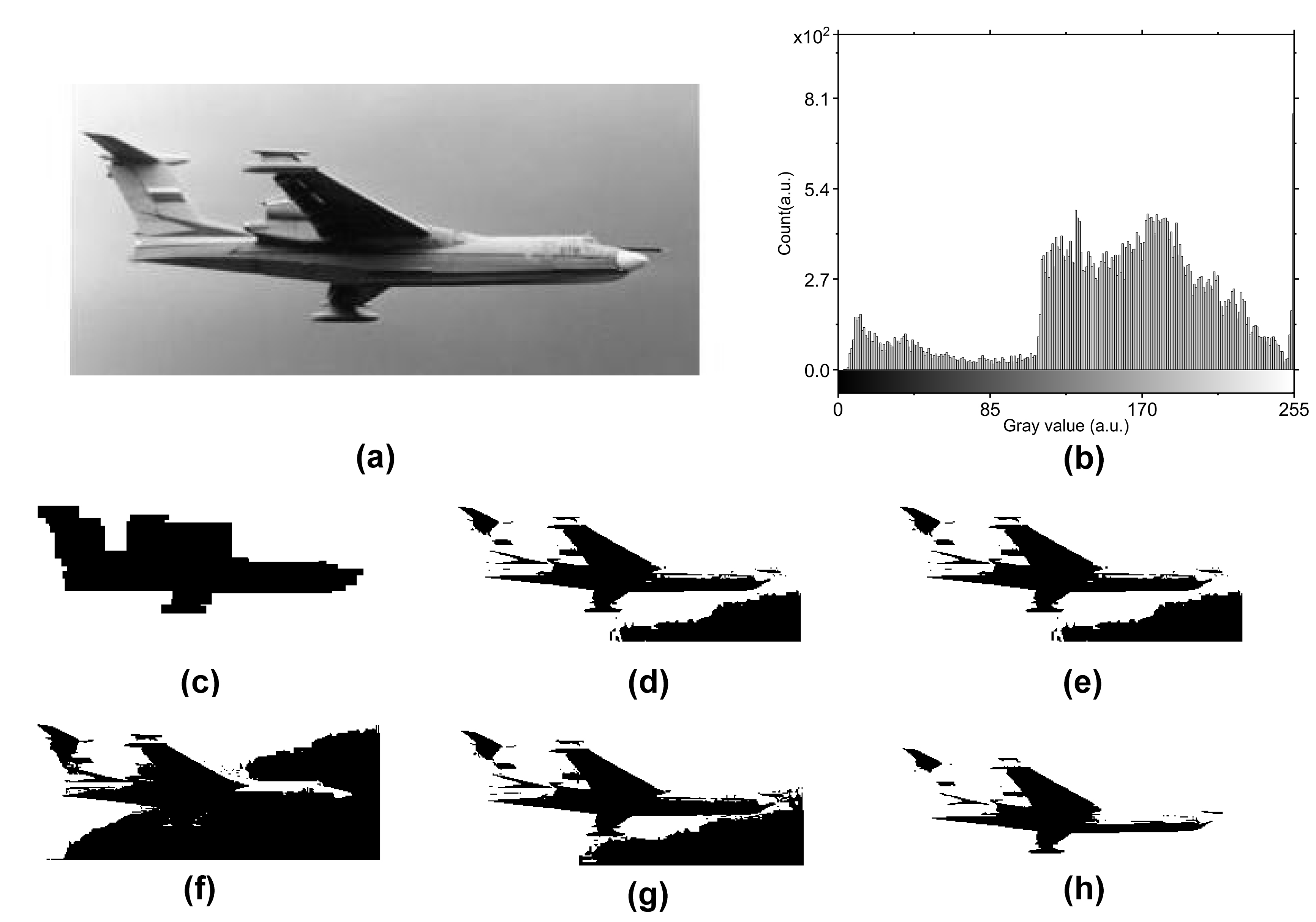}
	\caption{Segmentation results of airplane. (a) Original image. (b) Histogram of original image. (c) Ground truth. (d) Otsu. (e) K-mean. (f) SPCNN. (g) CCNN. (h) RCNN}
\end{figure}
\subsection{ Image fusion}
An x-ray differential imaging dataset of frog bones was used to evaluate the performance of RCNN in image fusion. This imaging technique output three channels simultaneously in a single scan, namely AC, DPC, and DFC channels \cite{RN72,RN73}. These three channels contain complementary information, presenting different structures of the sample. Image fusion in this imaging technique aims to fuse images from these three channels into one image, presenting all information compactly and efficiently. The fusion processes of the RCNN-based method are as follows:
\begin{enumerate}
	\item Decompose input images into directional sub-bands using nonsubsampled contourlet transform (NSCT);
	\item Process all sub-bands using RCNN, generating ignition maps for every sub-bands;
	\item Select pixels based on ignition results, composing a fused sub-band group;
	\item Reconstruct the fused sub-band group using inverse NSCT.
\end{enumerate}

Detailed pixel selection and sub-band fusion algorithms are given in the Supplementary material C. The fusion results of the RCNN-based method are compared with the other four common image fusion methods: wavelet \cite{RN74}, NSCT \cite{RN75}, NSCT-PCNN \cite{RN76}, and CCNN \cite{RN20}. Detailed parameter settings for these fusion algorithms can be found in the Supplementary material D. The experimental results are shown in Fig. 10 and Fig. 11.

As shown in Fig. 10, the high intensities at the extremity and bone cortex in the DFC and DPC channels are successfully transformed into the fused image by the RCNN-based algorithm. The meshwork structure of the bone trabecula in the DPC channel is well preserved. The heterogeneous bone marrow intensity can be clearly observed in the fusion results. The soft tissue that can be clearly observed in the DPC channel is fused into the output of the RCNN-based algorithm.

The comparison between the fusion results of different methods is shown in Fig. 11. The details of bone trabecula are not well preserved by the wavelet method in Fig. 11a and Fig. 11f. In the fusion results of the NSCT method, high intensities at extremity are missed in Fig. 11b. There are artifacts in the bone trabecula parts in Fig. 11g. Concerning the NSCT-PCNN method, strong noise appeared at the extremity and bone cortex parts in fused images. Moreover, details at the bone marrow and soft tissue parts are slightly distorted in Fig. 11d and Fig. 10i, with CCNN method. Generally, the fusion results of the RCNN-based method outperform the others in subjective evaluation.
\begin{figure}[t]
	\centering
	\includegraphics[width=0.5\textwidth]{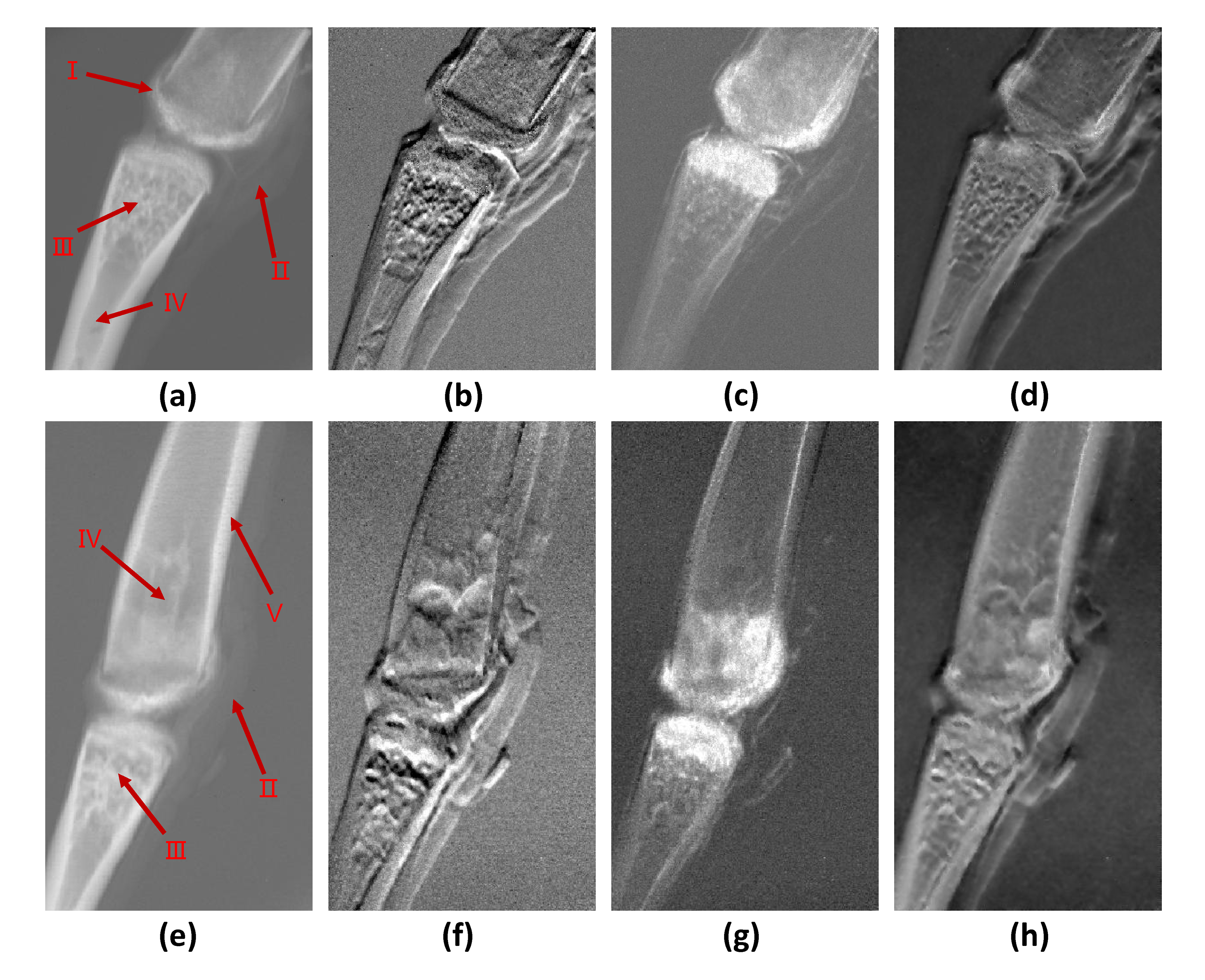}
	\caption{X-ray differential imaging fusion. (a) (e) Images from the AC channel. X-ray differential phase contrast imaging results of frog toes with extremity (I), soft tissue (II), bone trabecula (III), bone marrow (IV), and bone cortex (V). (b) (f) Images from the DPC channel. (c) (g) Images from the DFC channel. (d) (h) Fused images using RCNN-based algorithm.}
\end{figure}
\begin{figure}[t]
	\centering
	\includegraphics[width=0.5\textwidth]{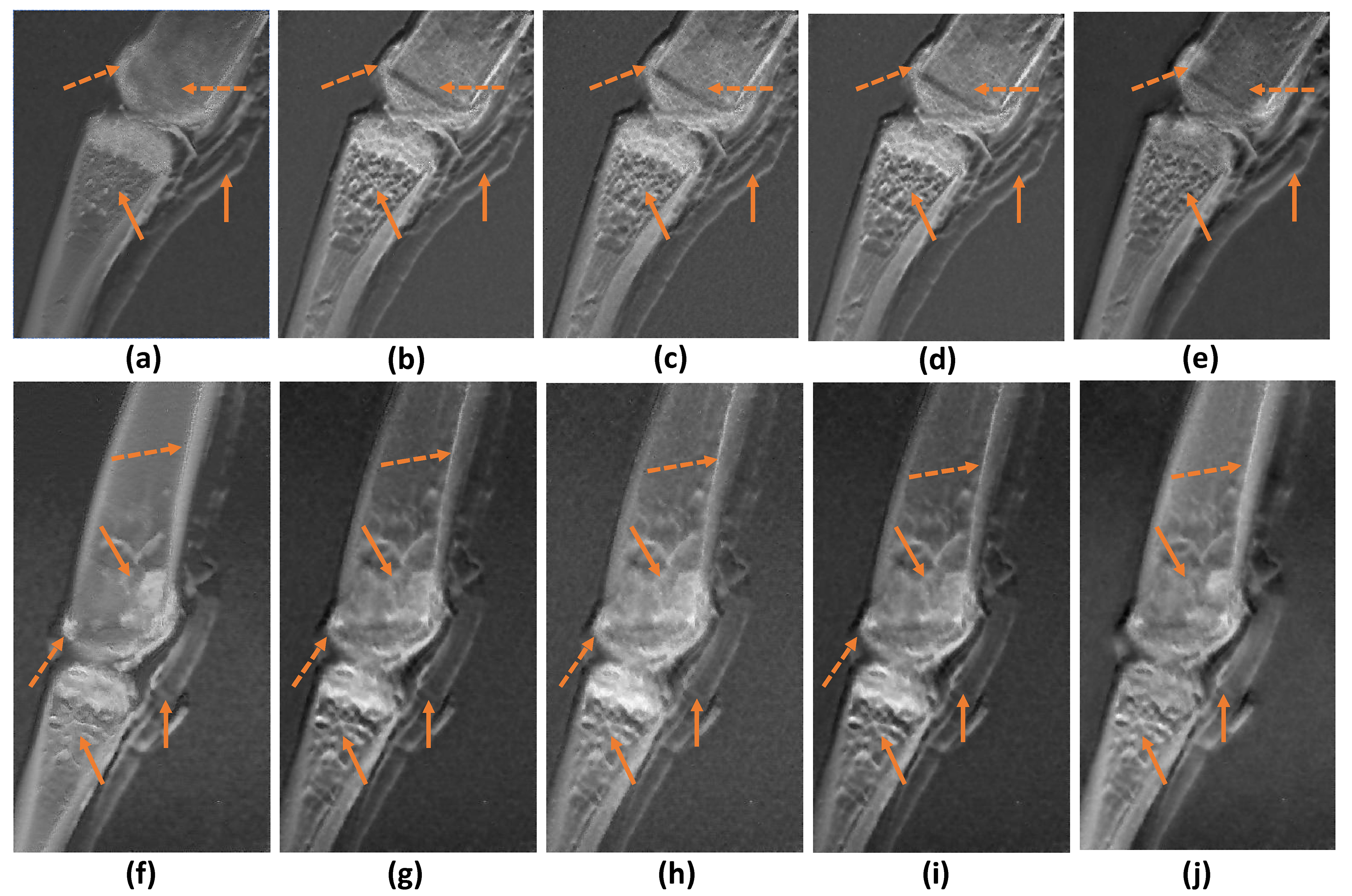}
	\caption{Fusion results. (a) (f) Wavelet. (b) (g) NSCT. (c) (h) NCST-PCNN. (d) (i) NSCT-CCNN. (e) (j) NSCT-RCNN. }
\end{figure}
\begin{figure}[t]
	\centering
	\includegraphics[width=0.5\textwidth]{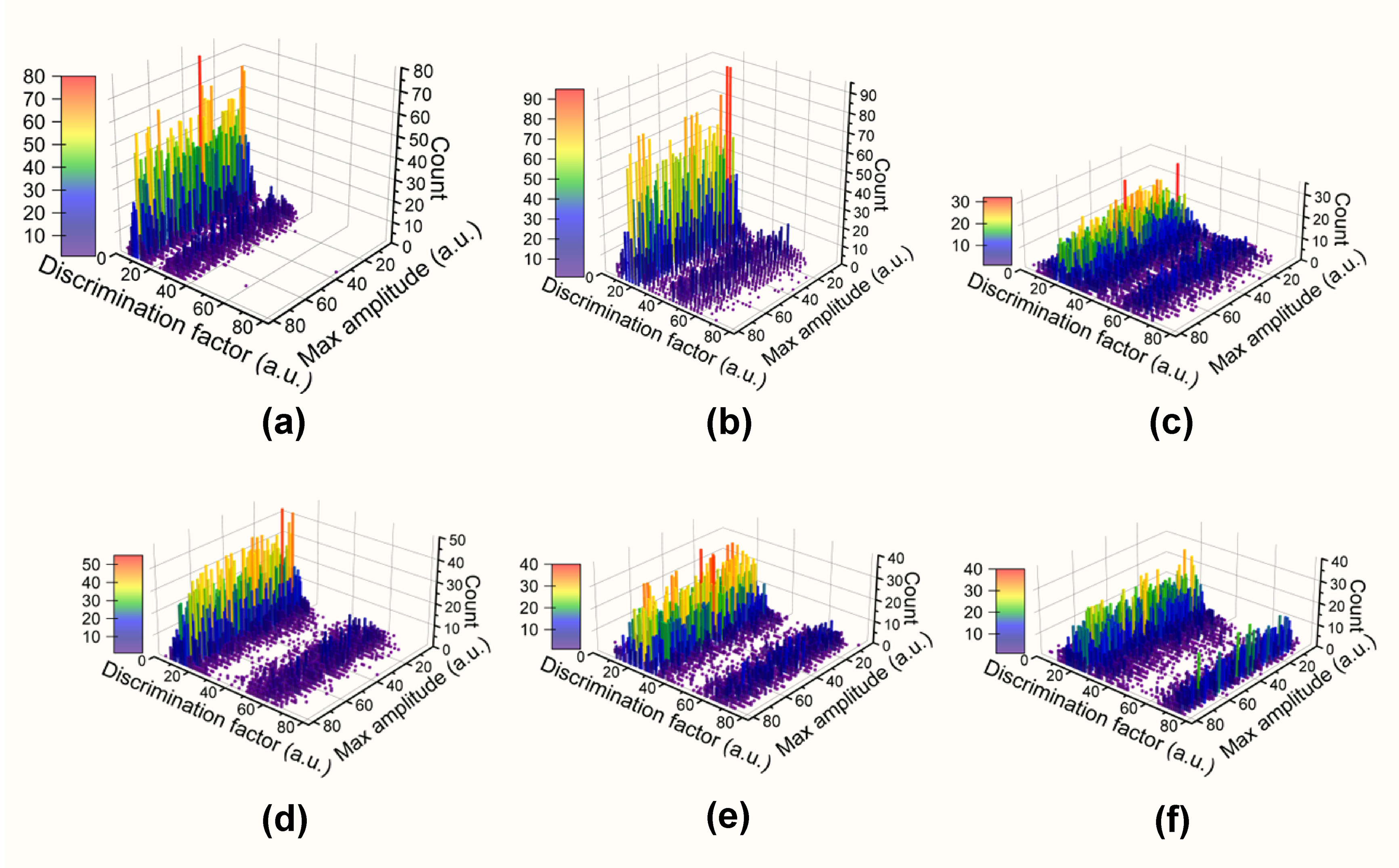}
	\caption{Three-dimensional histograms of the discrimination results. (a) Results of charge comparison. (b) Results of zero crossing. (c) Results of falling edge percentage slop. (d) Results of PCNN. (e) Results of ladder gradient. (f) Results of RCNN. }
\end{figure}
Additionally, several evaluation criteria are used to measure the results of all methods to evaluate the fusion performance objectively. These evaluation criteria \cite{RN77} are edge strength (ES), spatial frequency (SF), standard deviation (SD), entropy (H), feature mutual information (FMI), feature similarity index measure (FSIM), fusion factor (FF), and structural similarity index measure (SSIM). Detailed mathematical expressions of these evaluation criteria are given in the Supplementary material E. The evaluation results are given in Table 2. The best performance of each criterion is marked in bold.

From the objective evaluation results, the RCNN-based method outperforms others with regard to H and FF. The ES, SF, and FMI of it are very close to that of the other methods. These criteria denote the amount of information and details in the final fusion results. A good performance in these criteria indicates that RCNN successfully extracts information from the input images and presents it in the fused image with low noise and few artifacts. Regarding the last two criteria, SSIM and FSIM, the RCNN-based fusion algorithm notably  
lags other methods. However, this result is reasonable because these two criteria measure the similarity between the fused image and input images. If the details of all three input images are successfully transformed into the fused image, the similarity between the fused image and all three inputs will inevitably decrease. 

In conclusion, RCNN exhibits a state-of-the-art performance in image fusion applications. It can efficiently extract information from images while resisting the negative influence of noise. The fusion results of the RCNN-based method have sufficient information, fine details, low noise, and few artifacts.
\subsection{Pulse shape discrimination}
A data set of approximately ten thousand neutron and gamma-ray mixed pulse signals is used for the experiment. The discrimination processes of the RCNN-based method are as follows:
\begin{enumerate}
	\item Radiation pulse signals are processed by RCNN, generating an ignition map for each pulse signal;
	\item Parts of ignition maps that contain the information on pulse signals’ falling edge and delayed fluorescence are integrated;
	\item Use the integrated result of each pulse signal as the discrimination factor;
	\item Draw a histogram of discrimination factors. In the histogram, groups of neutron counts and gamma-ray counts are separated. The discrimination process is complete.
\end{enumerate}
Detailed discrimination parameters and information about the experimental setups for recording neutron and gamma-ray superposed field are given in the Supplementary material F. Moreover, the discrimination performance of the RCNN-based method is compared to other common discrimination algorithms, which are charge comparison (CC) \cite{RN62}, zero crossing (ZC) \cite{RN63}, falling edge percentage slop (FEPS) \cite{RN78}, PCNN \cite{RN66}, and ladder gradient (LG) \cite{RN67}. Parameter settings for these discrimination methods are given in the Supplementary material G, and the experimental results are shown in Fig. 12 and Table 3. It is worth noting that filtering is an important step in pulse shape discrimination, but discrimination methods exhibit different performances when coupled with various filtering methods. Consequently, all discrimination methods uniformly use the Fourier filter to pre-process radiation pulse signals in this study.

\begin{table*}
	\centering
	\caption{Objective Evaluation of Image Segmentation}
	\label{tab_demo}

	\resizebox{\linewidth}{!}{
		\tiny
		\begin{tabular}{ccccccccccc}
			\hline
			\hspace{-1cm}\textbf{Criteria} & \textbf{Otsu} & \textbf{K-mean} & \textbf{SPCNN} & \textbf{CCNN} & \textbf{RCNN}\\
			\hline
			\hspace{2cm}Fig. 6 flower\\
			\hspace{-1cm}\textbf{PA}   & \textbf{0.8600} & \textbf{0.8600}  & 0.7900& 0.8585  & 0.8342\\
			\hspace{-1cm}\textbf{IoU } & 0.4592 & 0.4592 & 0.5195 & 0.4505 & \textbf{0.5742} \\
			\hspace{-1cm}\textbf{Dice}  & 0.6294 & 0.6294   & 0.6837 & 0.6211  & \textbf{0.7295} \\
			\hspace{2cm}Fig. 7 CT\\
			\hspace{-1cm}\textbf{PA}   & 0.9577 & 0.9577  & 0.9691& 0.9532  & \textbf{0.9704}\\
			\hspace{-1cm}\textbf{IoU}  & 0.7528 & 0.7528 & \textbf{0.8325 }& 0.7255 & \textbf{0.8325}\\
			\hspace{-1cm}\textbf{Dice}  & 0.8590 & 0.8590   & \textbf{0.9086} & 0.8409  & \textbf{0.9086} \\
			\hspace{2cm}Fig. 8 apples\\
			\hspace{-1cm}\textbf{PA}   & 0.8166 & 0.8166  & 0.3843 & 0.7838  & \textbf{0.8515}\\
			\hspace{-1cm}\textbf{IoU}  & 0.7963 & 0.7963 & 0.3000 & 0.7593 & \textbf{0.8372}\\
			\hspace{-1cm}\textbf{Dice}  & 0.8866	& 0.8866	&0.4615 &0.8632	&\textbf{0.9114 }\\
			\hline
		\end{tabular}
	}
\end{table*}
\begin{table*}
	\centering
	\caption{Objective Evaluation of Image Fusion}
	\label{tab_demo1}

	\resizebox{\linewidth}{!}{
		
		\begin{tabular}{ccccccccccc}
			\hline
			\textbf{Criteria}	&\textbf{Wavelet}	&\textbf{NSCT}	&\textbf{NSCT-PCNN}	&N\textbf{SCT-CCNN}	&\textbf{NSCT-RCNN}\\
			\hline
			Objective evaluation of Fig. 11a - Fig. 11e\\
			
			\textbf{ES }	&0.1236	&\textbf{0.2513}	&0.2352	&\textbf{0.2513}	&0.2430\\
			\textbf{H}	&6.0068	&6.2220	&6.3700	&6.2220	&\textbf{6.3954}\\
			\textbf{SD}	&0.1001	&0.1248	&\textbf{0.1267}	&0.1248	&0.1123\\
			\textbf{SF}	&\textbf{8.8725}	&6.9122	&7.6678	&6.9120	&7.2361\\
			\textbf{FMI}	&0.8246	&\textbf{0.8648}	&0.8470	&\textbf{0.8648	}&0.8601\\
			\textbf{FF}	&12.852	&13.070	&13.004	&13.070	&\textbf{13.536}\\
			\textbf{SSIM}	&0.9958	&0.9958	&\textbf{0.9959	}&0.9958	&0.9953\\
			\textbf{FSIM}	&0.9323	&0.9401	&\textbf{0.9408}	&0.9401	&0.9302\\
			Objective evaluation of Fig. 11f - Fig. 11j\\
			\textbf{ES}	&0.1350	&\textbf{0.2974}	&0.2727	&0.2896	&0.2834\\
			\textbf{H}	&6.7138	&6.5689	&6.6733	&6.7252	&\textbf{6.9636}\\
			\textbf{SD}	&0.1422	&0.1386	&0.1420	&\textbf{0.1531}	&0.1498\\
			\textbf{SF}	&\textbf{9.7547}	&5.5169	&6.4207	&5.2795	&6.9638\\
			\textbf{FMI}	&0.8376	&\textbf{0.8742}	&0.8594	&0.8697	&0.8729\\
			\textbf{FF}	&13.740	&13.633	&13.562	&13.6264	&\textbf{14.431}\\
			\textbf{SSIM}	&0.9960	&0.9946	&\textbf{0.9969}	&0.9958	&0.9960\\
			\textbf{FSIM}	&0.9136	&\textbf{0.9403}	&0.9395	&\textbf{0.9403}	&0.9376\\
			\hline
		\end{tabular}
	}
\end{table*}
\begin{table*}
	\centering
	\caption{Objective Evaluation of Pulse Shape Discrimination}
	\label{tab_demo2}

	\resizebox{\linewidth}{!}{
		
		\begin{tabular}{ccccccccccc}
			\hline
			\textbf{Criteria}	&\textbf{CC}	&\textbf{ZC}	&\textbf{FEPS}	&\textbf{PCNN}	&\textbf{LG}	&\textbf{RCNN}\\
			\hline
			\textbf{FOM}	&1.31±0.01	&1.10±0.01	&1.22±0.01	&2.00±0.01	&1.47±0.01	&\textbf{2.28}±0.01\\
			\textbf{Time consumption (s)}	&1.30±0.01	&\textbf{0.40}±0.01	&0.57±0.01	&3.35±0.01	&1.51±0.01	&2.72±0.01\\
			\hline
		\end{tabular}
	}
\end{table*}

As shown in Fig. 12, each histogram has two groups with a Gaussian distribution (group of gamma-ray counts on the left and neutron counts on the right side). If both groups exhibit Gaussian distributions with small variances and separate significantly with a clear gap, the discrimination process is successful. The CC method has a good result with a clean separation between the neutron and gamma-ray groups, as shown in Fig. 12a. However, traditional fast discrimination methods perform poorly, as shown in Fig. 12b and 11c. The variance of their groups’ Gaussian distribution is large, and many pulse signal counts locate between two groups. In Fig. 12d and 12e, the discrimination is successful with good groups that have Gaussian distribution with low variance and clean separation between groups. This attributes to the outstanding feature extraction ability of the PCNN and PCNN’s derived models. Finally, the performance of the RCNN-based method is extraordinary, with a clean gap between groups and a surprisingly low variance of the neutron group.

Table 3 shows the objective evaluation results of discrimination performance. The best performance of each criterion is marked in bold. The evaluation criteria are FOM-value and time (CPU processing time for discrimination of all pulse signals). The FOM-value is a standard measurement for discrimination performance in the pulse shape discrimination field. A larger FOM-value indicates better discrimination performance. The calculation method for FOM-value is given in the Supplementary material H. The CC, PCNN, and RCNN methods calculate discrimination factors via integration process, which is more accurate than methods based on gradient calculation (ZC, FEPS, and LG). However, the time consumption of integration-based methods is higher than that of gradient-based methods.

From the results presented in Table 3, the PCNN, LG, and RCNN methods perform better than traditional gradient-based methods (ZC and FEPS), and traditional integration-based method (CC). These results come from the excellent information extraction and analyzing capabilities of PCNN-derived models. Additionally, the RCNN-based discrimination method significantly outperforms others, with the highest FOM-value. Its time consumption is less than that of PCNN, which bases on the same kinds of feature extraction and integration-based discrimination factor calculation.

In conclusion, the RCNN-based discrimination method significantly outperforms other algorithms mentioned above. This outstanding performance of RCNN attributes to its better accuracy and low computational complexity, properties that benefit from the random inactivation weight matrix of link input.

\section{Conclusion and future work}
The PCNN and its derived models played a vital role in neuromorphic computing and third-generation neural networks. This study analyzed the difficulties that curtail the PCNN-derived models from achieving characteristics of brain-like computing. Three significant distinctions between the PCNN and the human brain were listed: limited neural connection, high computational cost, and lack of stochastic property. This study, therefore, proposed the random-coupled neural network (RCNN) that simultaneously solves these drawbacks of PCNN-derived models. RCNN considers the link input of neurons as a stochastic process, in which a central neuron randomly receives stimulus from its surrounding neurons. This property is achieved by a stochastic process called random inactivation. Neurons in the RCNN architecture are connected through the random inactivation matrix of link input, which randomly closes some connections between neurons. This stochastic inactivation releases the computational burden of neural link input, making a vast neural connection possible. In the RCNN model, a central neuron is connected to more than 20 neurons, and this number can be expanded up to 400. Consequently, stochastic properties, together with low computational complexity and vast neural connection, make RCNN achieve high accuracy, strong anti-noise ability, and, most importantly, biological neural response. This paper validated the image and video processing properties of RCNN, finding that RCNN encodes constant stimuli as periodic spike trains and periodic stimuli as chaotic spike trains. This neural response is the same as that of biological neural systems.

Moreover, this study conducted experiments to explore RCNN’s performance in two-dimensional and one-dimensional signal processing tasks. Experiments were divided into image segmentation, image fusion, and pulse shape discrimination. In image segmentation, the RCNN-based method significantly outperforms others, with segmentation results that exhibit little noise, clear boundaries, and better objective criteria performance. It can segment images commonly operable only by hierarchical methods into the front and background layers. Its image feature analysis ability demonstrates its potential to aid machine learning methods in semantic segmentation tasks. In image fusion, the RCNN-based method efficiently extracts information from images while resisting the negative influence of noise. The fusion results have sufficient information, fine details, low noise, and few artifacts. In pulse shape discrimination, the RCNN-based method significantly outperforms traditional methods and even state-of-the-art methods based on PCNN-derived models. Its discrimination results exhibit Gaussian distribution with low variance for both neutron and gamma-ray groups and a clear gap between them. Although an improvement in PCNN’s neuromorphic computing has been made in this study, the applications that RCNN can be capable of are far from complete. Further research on RCNN’s architecture is still needed. Here a few possible future research directions are addressed:
\begin{enumerate}
	\item In this study, RCNN’s video processing mechanism was researched. It was demonstrated that the periodic-chaotic spike trains encoding property make object recognition in videos possible without feature extraction. Video processing applications should be an essential research topic in the future;
	\item In this study, RCNN’s parameters are selected with experience and extensive experiments. An automatic parameter decision method for RCNN could play a vital role in RCNN’s applications in numerous fields in the future.
	\item In this study, RCNN has been validated as a powerful feature extractor. It should be applied to various signal processing tasks such as semantic image segmentation, moving object tracking, and object recognition.
\end{enumerate}
\section{Acknowledgment}
The authors thank Dr. Guibin Zan from SLAC National Accelerator Laboratory, Stanford Synchrotron Radiation Lightsource, for valuable discussions and technical support.


\begin{thebibliography}{99}
	
	\bibitem{RN1}
	E. J. O’Gorman, "Machine learning ecological networks," \textit{Science}, vol. 377, no. 6609, pp. 918-919, 2022, doi: 10.1126/science.add7563.
	
	\bibitem{RN2}
	D. Silver et al., "Mastering the game of Go without human knowledge," \textit{Nature}, vol. 550, no. 7676, pp. 354-359, 2017/10/01 2017, doi: 10.1038/nature24270.
	
	\bibitem{RN3}
	K. Roy, A. Jaiswal, and P. Panda, "Towards spike-based machine intelligence with neuromorphic computing," \textit{Nature}, vol. 575, no. 7784, pp. 607-617, 2019/11/01 2019, doi: 10.1038/s41586-019-1677-2.
	\bibitem{RN4}
	W. Maass, "Networks of spiking neurons: The third generation of neural network models," \textit{Neural Networks}, vol. 10, no. 9, pp. 1659-1671, Dec 1997, doi: 10.1016/s0893-6080(97)00011-7.
	
	\bibitem{RN5}
	W. S. McCulloch and W. Pitts, "A logical calculus of the ideas immanent in nervous activity. 1943," \textit{Bulletin of mathematical biology}, vol. 52, no. 1-2, pp. 99-115; discussion 73-97, 1990 1990, doi: 10.1016/s0092-8240(05)80006-0.
	
	\bibitem{RN6}
	E. M. Izhikevich, "Simple model of spiking neurons," \textit{IEEE Transactions on Neural Networks}, vol. 14, no. 6, pp. 1569-1572, Nov 2003, doi: 10.1109/tnn.2003.820440.
	
	\bibitem{RN7}
	M. Liu, F. Zhao, X. Jiang, H. Zhang, and H. Zhou, "Parallel binary image cryptosystem via spiking neural networks variants," \textit{International Journal of Neural Systems}, p. 2150014, 2021, doi: 10.1142/S0129065721500143.
	
	\bibitem{RN8}
	W. Olin-Ammentorp, K. Beckmann, C. D. Schuman, J. S. Plank, and N. C. Cady, "Stochasticity and robustness in spiking neural networks," \textit{Neurocomputing}, vol. 419, pp. 23-36, 2021/01/02/ 2021, doi: 10.1016/j.neucom.2020.07.105.
	
	\bibitem{RN9}
	J. L. Johnson, "Pulse-coupled neural nets: translation, rotation, scale, distortion, and intensity signal invariance for images," \textit{Appl. Opt.}, vol. 33, no. 26, pp. 6239-6253, 1994/09/10 1994, doi: 10.1364/AO.33.006239.
	
	\bibitem{RN10}
	H. Jia, Z. Xing, and W. Song, "Three Dimensional Pulse Coupled Neural Network Based on Hybrid Optimization Algorithm for Oil Pollution Image Segmentation," \textit{Remote Sensing}, vol. 11, no. 9, 2019, doi: 10.3390/rs11091046.
	\bibitem{RN11}
	K. He, R. Wang, D. Tao, J. Cheng, and W. Liu, "Color Transfer Pulse-Coupled Neural Networks for Underwater Robotic Visual Systems," \textit{IEEE Access}, vol. 6, pp. 32850-32860, 2018, doi: 10.1109/ACCESS.2018.2845855.
	
	\bibitem{RN12}
	R. Shanker and M. Bhattacharya, "Automated Diagnosis system for detection of the pathological brain using Fast version of Simplified Pulse-Coupled Neural Network and Twin Support Vector Machine," \textit{Multimedia Tools and Applications}, vol. 80, no. 20, pp. 30479-30502, 2021/08/01 2021, doi: 10.1007/s11042-021-10937-6.
	
	\bibitem{RN13}
	M. M. Altaf, "A hybrid deep learning model for breast cancer diagnosis based on transfer learning and pulse-coupled neural networks," \textit{Mathematical Biosciences and Engineering}, vol. 18, no. 5, pp. 5029-5046, 2021, doi: 10.3934/mbe.2020392.
	
	\bibitem{RN14}
	L. Li and H. Ma, "Pulse Coupled Neural Network-Based Multimodal Medical Image Fusion via Guided Filtering and WSEML in NSCT Domain," \textit{Entropy}, vol. 23, no. 5, p. 591, 2021, doi: 10.3390/e23050591.
	
	\bibitem{RN15}
	K. K. Thyagharajan and G. Kalaiarasi, "Pulse coupled neural network based near-duplicate detection of images (PCNN–NDD)," \textit{Advances in Electrical and Computer Engineering}, vol. 18, no. 3, pp. 87-97, 2018, doi: 10.4316/AECE.2018.03012.
	\bibitem{RN16}
	H. Liu, Z. Zuo, P. Li, B. Liu, L. Chang, and Y. Yan, "Anti-noise performance of the pulse coupled neural network applied in discrimination of neutron and gamma-ray," \textit{Nuclear Science and Techniques}, vol. 33, no. 6, p. 75, 2022/07/08 2022, doi: 10.1007/s41365-022-01054-6.
	
	\bibitem{RN17}
	J. Lian et al., "An Overview of Image Segmentation Based on Pulse-Coupled Neural Network," \textit{Archives of Computational Methods in Engineering}, vol. 28, no. 2, pp. 387-403, 2021/03/01 2021, doi: 10.1007/s11831-019-09381-5.
	
	\bibitem{RN18}
	H. Liu, M. Liu, D. Li, W. Zheng, L. Yin, and R. Wang, "Recent Advances in Pulse-Coupled Neural Networks with Applications in Image Processing," \textit{Electronics}, vol. 11, no. 20, 2022, doi: 10.3390/electronics11203264.
	
	\bibitem{RN19}
	R. M. Siegel, "Non-linear dynamical system theory and primary visual cortical processing," \textit{Physica D: Nonlinear Phenomena}, vol. 42, no. 1, pp. 385-395, 1990/06/01/ 1990, doi: 10.1016/0167-2789(90)90090-C.
	
	\bibitem{RN20}
	J. Liu, J. Lian, J. C. Sprott, Q. Liu, and Y. Ma, "The Butterfly Effect in Primary Visual Cortex," \textit{IEEE Transactions on Computers}, pp. 1-1, 2022, doi: 10.1109/TC.2022.3173080.
	
	\bibitem{RN21}
	R. Eckhorn, H. J. Reitboeck, M. Arndt, and P. Dicke, "Feature Linking via Synchronization among Distributed Assemblies: Simulations of Results from Cat Visual Cortex," \textit{Neural Computation}, vol. 2, no. 3, pp. 293-307, 1990, doi: 10.1162/neco.1990.2.3.293.
	
	\bibitem{RN22}
	S. Thorpe and J. Gautrais, "Rapid visual processing using spike asynchrony," \textit{Advances in neural information processing systems}, vol. 9, 1996.
	
	\bibitem{RN23}
	T. Natschläger and B. Ruf, "Spatial and temporal pattern analysis via spiking neurons," \textit{Network: Computation in Neural Systems}, vol. 9, no. 3, p. 319, 1998/08/01 1998, doi: 10.1088/0954-898X/9/3/003.
	
	\bibitem{RN24}
	Y. Chen, S. K. Park, Y. Ma, and R. Ala, "A New Automatic Parameter Setting Method of a Simplified PCNN for Image Segmentation," \textit{IEEE Transactions on Neural Networks}, vol. 22, no. 6, pp. 880-892, 2011, doi: 10.1109/TNN.2011.2128880.
	
	\bibitem{RN25}
	K. Zhan, H. Zhang, and Y. Ma, "New Spiking Cortical Model for Invariant Texture Retrieval and Image Processing," \textit{IEEE Transactions on Neural Networks}, vol. 20, no. 12, pp. 1980-1986, 2009, doi: 10.1109/TNN.2009.2030585.
	\bibitem{RN26}
	Y. Chen, Y. Ma, D. H. Kim, and S. K. Park, "Region-Based Object Recognition by Color Segmentation Using a Simplified PCNN," \textit{IEEE Transactions on Neural Networks and Learning Systems}, vol. 26, no. 8, pp. 1682-1697, 2015, doi: 10.1109/TNNLS.2014.2351418.
	
	\bibitem{RN27}
	Y. Ma, L. Wang, Y. Ma, M. Dong, S. Du, and X. Sun, "An SPCNN-GVF-based approach for the automatic segmentation of left ventricle in cardiac cine MR images," \textit{International Journal of Computer Assisted Radiology and Surgery}, vol. 11, no. 11, pp. 1951-1964, 2016/11/01 2016, doi: 10.1007/s11548-016-1429-9.
	
	\bibitem{RN28}
	L. Zhang, G. Zeng, J. Wei, and Z. Xuan, "Multi-Modality Image Fusion in Adaptive-Parameters SPCNN Based on Inherent Characteristics of Image," \textit{IEEE Sensors Journal}, vol. 20, no. 20, pp. 11820-11827, 2020, doi: 10.1109/JSEN.2019.2948783.
	
	\bibitem{RN29}
	Q. Liu, Z. Liu, S. Yong, K. Jia, and N. Razmjooy, "Computer-aided breast cancer diagnosis based on image segmentation and interval analysis," \textit{Automatika}, vol. 61, no. 3, pp. 496-506, 2020/07/02 2020, doi: 10.1080/00051144.2020.1785784.
	
	\bibitem{RN30}
	M. Chaturvedi, M. Kaur, N. Rakesh, and P. Nand, "Object Recognition Using Image Segmentation," in \textit{2020 Sixth International Conference on Parallel, Distributed and Grid Computing (PDGC)}, 6-8 Nov. 2020, pp. 550-556, doi: 10.1109/PDGC50313.2020.9315803.
	
	\bibitem{RN31}
	M. Shi, S. Jiang, H. Wang, and B. Xu, "A Simplified pulse-coupled neural network for adaptive segmentation of fabric defects," \textit{Machine Vision and Applications}, vol. 20, no. 2, pp. 131-138, 2009/02/01 2009, doi: 10.1007/s00138-007-0113-z.
	
	\bibitem{RN32}
	Z. Li, Y. Liu, R. Walker, R. Hayward, and J. Zhang, "Towards automatic power line detection for a UAV surveillance system using pulse coupled neural filter and an improved Hough transform," \textit{Machine Vision and Applications}, vol. 21, no. 5, pp. 677-686, 2010/08/01 2010, doi: 10.1007/s00138-009-0206-y.
	
	\bibitem{RN33}
	N. Yang, H. Chen, Y. Li, and X. Hao, "Coupled Parameter Optimization of PCNN Model and Vehicle Image Segmentation," \textit{Journal of Transportation Systems Engineering and Information Technology}, vol. 12, no. 1, pp. 48-54, 2012/02/01 2012, doi: 10.1016/S1570-6672(11)60184-0.
	
	\bibitem{RN34}
	H. Li, X. Jin, N. Yang, and Z. Yang, "The recognition of landed aircrafts based on PCNN model and affine moment invariants," \textit{Pattern Recognition Letters}, vol. 51, pp. 23-29, 2015/01/01 2015, doi: 10.1016/j.patrec.2014.07.021.
	
	\bibitem{RN35}
	D. Im, D. Han, S. Choi, S. Kang, and H. J. Yoo, "DT-CNN: Dilated and Transposed Convolution Neural Network Accelerator for Real-Time Image Segmentation on Mobile Devices," in \textit{2019 IEEE International Symposium on Circuits and Systems (ISCAS)}, 26-29 May 2019, pp. 1-5, doi: 10.1109/ISCAS.2019.8702243.
	\bibitem{RN36}
	Y. Ma, R. Dai, L. Li, and L. Wei, "Image segmentation of embryonic plant cell using pulse-coupled neural networks," \textit{Chinese Science Bulletin}, vol. 47, no. 2, pp. 169-173, 2002/01/01 2002, doi: 10.1360/02tb9040.
	
	\bibitem{RN37}
	Y. Lu, J. Miao, L. Duan, Y. Qiao, and R. Jia, "A new approach to image segmentation based on simplified region growing PCNN," \textit{Applied Mathematics and Computation}, vol. 205, no. 2, pp. 807-814, 2008/11/15 2008, doi: 10.1016/j.amc.2008.05.029.
	
	\bibitem{RN38}
	S. Wei, Q. Hong, and M. Hou, "Automatic image segmentation based on PCNN with adaptive threshold time constant," \textit{Neurocomputing}, vol. 74, no. 9, pp. 1485-1491, 2011/04/01 2011, doi: 10.1016/j.neucom.2011.01.005.
	
	\bibitem{RN39}
	N. Otsu, "A threshold selection method from gray-level histograms," \textit{IEEE Transactions on Systems, Man, and Cybernetics}, vol. 9, no. 1, pp. 62-66, 1979.
	
	\bibitem{RN40}
	J. T. Tou and R. C. Gonzalez, "Pattern recognition principles," 1974.
	
	\bibitem{RN41}
	P. Shan, "Image segmentation method based on K-mean algorithm," \textit{EURASIP Journal on Image and Video Processing}, vol. 2018, no. 1, p. 81, 2018/09/03 2018, doi: 10.1186/s13640-018-0322-6.
	
	\bibitem{RN42}
	D. Zhou, H. Zhou, C. Gao, and Y. Guo, "Simplified parameters model of PCNN and its application to image segmentation," \textit{Pattern Analysis and Applications}, vol. 19, no. 4, pp. 939-951, 2016/11/01 2016, doi: 10.1007/s10044-015-0462-6.
	
	\bibitem{RN43}
	B. Huang, F. Yang, M. Yin, X. Mo, and C. Zhong, "A Review of Multimodal Medical Image Fusion Techniques," \textit{Computational and Mathematical Methods in Medicine}, vol. 2020, p. 8279342, 2020/04/23 2020, doi: 10.1155/2020/8279342.
	
	\bibitem{RN44}
	H. Zhang, H. Xu, X. Tian, J. Jiang, and J. Ma, "Image fusion meets deep learning: A survey and perspective," \textit{Information Fusion}, vol. 76, pp. 323-336, 2021/12/01 2021, doi: 10.1016/j.inffus.2021.06.008.
	
	\bibitem{RN45}
	G. Li, Y. Lin, and X. Qu, "An infrared and visible image fusion method based on multi-scale transformation and norm optimization," \textit{Information Fusion}, vol. 71, pp. 109-129, 2021/07/01 2021, doi: 10.1016/j.inffus.2021.02.008.
	
	\bibitem{RN46}
	Y. Zhou et al., "A Survey of Multi-Focus Image Fusion Methods," \textit{Applied Sciences}, vol. 12, no. 12, p. 6281, 2022.
	
	\bibitem{RN47}
	R. Hou, D. Zhou, R. Nie, D. Liu, and X. Ruan, "Brain CT and MRI medical image fusion using convolutional neural networks and a dual-channel spiking cortical model," \textit{Medical \& Biological Engineering \& Computing}, vol. 57, no. 4, pp. 887-900, 2019/04/01 2019, doi: 10.1007/s11517-018-1935-8.
	
	\bibitem{RN48}
	E. Coello et al., "Fourier domain image fusion for differential X-ray phase-contrast breast imaging," \textit{European Journal of Radiology}, vol. 89, pp. 27-32, 2017/04/01 2017, doi: 10.1016/j.ejrad.2017.01.019.
	
	\bibitem{RN49}
	J. Chen, S. Paris, and F. Durand, "Real-time edge-aware image processing with the bilateral grid," \textit{ACM Transactions on Graphics}, vol. 26, no. 3, pp. 103-es, 2007, doi: 10.1145/1276377.1276506.
	
	\bibitem{RN50}
	R. Fattal, M. Agrawala, and S. Rusinkiewicz, "Multiscale shape and detail enhancement from multi-light image collections," \textit{ACM Transactions on Graphics}, vol. 26, no. 3, pp. 51-es, 2007, doi: 10.1145/1276377.1276441.
	
	\bibitem{RN51}
	M. Ehlers, "Spectral characteristics preserving image fusion based on Fourier domain filtering," in \textit{Remote Sensing for Environmental Monitoring, GIS Applications, and Geology IV}, 2004, vol. 5574, SPIE, pp. 1-13, doi: 10.1117/12.565160.
	
	\bibitem{RN52}
	G. Pajares and J. Manuel de la Cruz, "A wavelet-based image fusion tutorial," \textit{Pattern Recognition}, vol. 37, no. 9, pp. 1855-1872, 2004/09/01 2004, doi: 10.1016/j.patcog.2004.03.010.
	
	\bibitem{RN53}
	G. Bhatnagar, Q. M. J. Wu, and Z. Liu, "Directive Contrast Based Multimodal Medical Image Fusion in NSCT Domain," \textit{IEEE Transactions on Multimedia}, vol. 15, no. 5, pp. 1014-1024, 2013, doi: 10.1109/TMM.2013.2244870.
	
	\bibitem{RN54}
	Y. Yang, S. Tong, S. Huang, and P. Lin, "Multifocus Image Fusion Based on NSCT and Focused Area Detection," \textit{IEEE Sensors Journal}, vol. 15, no. 5, pp. 2824-2838, 2015, doi: 10.1109/JSEN.2014.2380153.
	
	\bibitem{RN55}
	P. Ganasala and V. Kumar, "Multimodality medical image fusion based on new features in NSST domain," \textit{Biomedical Engineering Letters}, vol. 4, no. 4, pp. 414-424, 2014/12/01 2014, doi: 10.1007/s13534-014-0161-z.
	
	\bibitem{RN56}
	P. Ganasala and V. Kumar, "Feature-Motivated Simplified Adaptive PCNN-Based Medical Image Fusion Algorithm in NSST Domain," \textit{Journal of Digital Imaging}, vol. 29, no. 1, pp. 73-85, 2016/02/01 2016, doi: 10.1007/s10278-015-9806-4.
	
	\bibitem{RN57}
	C. Panigrahy, A. Seal, and N. K. Mahato, "MRI and SPECT Image Fusion Using a Weighted Parameter Adaptive Dual Channel PCNN," \textit{IEEE Signal Processing Letters}, vol. 27, pp. 690-694, 2020, doi: 10.1109/LSP.2020.2989054.
	
	\bibitem{RN58}
	S. Ding, X. Zhao, H. Xu, Q. Zhu, and Y. Xue, "NSCT-PCNN image fusion based on image gradient motivation," \textit{IET Computer Vision}, vol. 12, no. 4, pp. 377-383, 2018, doi: 10.1049/iet-cvi.2017.0285.
	
	\bibitem{RN59}
	S. Cheng, M. Qiguang, and X. Pengfei, "A novel algorithm of remote sensing image fusion based on Shearlets and PCNN," \textit{Neurocomputing}, vol. 117, pp. 47-53, 2013/10/06 2013, doi: 10.1016/j.neucom.2012.10.025.
	
	\bibitem{RN60}
	D. Cester et al., "Pulse shape discrimination with fast digitizers," \textit{Nuclear Instruments and Methods in Physics Research Section A: Accelerators, Spectrometers, Detectors and Associated Equipment}, vol. 748, pp. 33-38, 2014/06/01 2014, doi: 10.1016/j.nima.2014.02.032.
	
	\bibitem{RN61}
	M. L. Roush, M. A. Wilson, and W. F. Hornyak, "Pulse shape discrimination," \textit{Nuclear Instruments and Methods}, vol. 31, no. 1, pp. 112-124, 1964/12/01 1964, doi: 10.1016/0029-554X(64)90333-7.
	
	\bibitem{RN62}
	D. Wolski, M. Moszyński, T. Ludziejewski, A. Johnson, W. Klamra, and Ö. Skeppstedt, "Comparison of $n-\gamma$ discrimination by zero-crossing and digital charge comparison methods," \textit{Nuclear Instruments and Methods in Physics Research Section A: Accelerators, Spectrometers, Detectors and Associated Equipment}, vol. 360, no. 3, pp. 584-592, 1995/06/15 1995, doi: 10.1016/0168-9002(95)00037-2.
	
	\bibitem{RN63}
	P. Sperr, H. Spieler, M. R. Maier, and D. Evers, "A simple pulse-shape discrimination circuit," \textit{Nuclear Instruments and Methods}, vol. 116, no. 1, pp. 55-59, 1974/03/15 1974, doi: 10.1016/0029-554X(74)90578-3.
	
	\bibitem{RN64}
	S. Pai, W. F. Piel, D. B. Fossan, and M. R. Maier, "A versatile electronic pulse-shape discriminator," \textit{Nuclear Instruments and Methods in Physics Research Section A: Accelerators, Spectrometers, Detectors and Associated Equipment}, vol. 278, no. 3, pp. 749-754, 1989/06/15 1989, doi: 10.1016/0168-9002(89)91199-6.
	
	\bibitem{RN65}
	M. Liu, B. Liu, Z. Zuo, L. Wang, G. Zan, and X. Tuo, "Toward a fractal spectrum approach for neutron and gamma pulse shape discrimination," \textit{Chinese Physics C}, vol. 40, no. 6, p. 066201, 2016, doi: 10.1088/1674-1137/40/6/066201.
	
	\bibitem{RN66}
	H. Liu, Y. Cheng, Z. Zuo, T. Sun, and K. Wang, "Discrimination of neutrons and gamma rays in plastic scintillator based on pulse-coupled neural network," \textit{Nuclear Science and Techniques}, vol. 32, no. 8, p. 82, 2021/08/05 2021, doi: 10.1007/s41365-021-00915-w.
	
	\bibitem{RN67}
	H. Liu, M. Liu, Y. Xiao, P. Li, Z. Zuo, and Y. Zhan, "Discrimination of neutron and gamma ray using the ladder gradient method and analysis of filter adaptability," \textit{Nuclear Science and Techniques}, vol. 33, no. 12, p. 159, 2022/12/09 2022, doi: 10.1007/s41365-022-01136-5.
	
	\bibitem{RN68}
	B. Albertina et al., "Radiology data from the cancer genome atlas lung adenocarcinoma [tcga-luad] collection," \textit{The Cancer Imaging Archive}, 2016, doi: 10.7937/K9/TCIA.2016.JGNIHEP5.
	
	\bibitem{RN69}
	F.-F. Li, R. Fergus, and P. Perona, "One-shot learning of object categories," \textit{IEEE Transactions on Pattern Analysis and Machine Intelligence}, vol. 28, no. 4, pp. 594-611, 2006, doi: 10.1109/TPAMI.2006.79.
	
	\bibitem{RN70}
	J. N. Kapur, P. K. Sahoo, and A. K. C. Wong, "A new method for gray-level picture thresholding using the entropy of the histogram," \textit{Computer Vision, Graphics, and Image Processing}, vol. 29, no. 3, pp. 273-285, 1985/03/01 1985, doi: 10.1016/0734-189X(85)90125-2.
	
	\bibitem{RN71}
	Z. Krawczyk and J. Starzyński, "Segmentation of bone structures with the use of deep learning techniques," \textit{Bulletin of the Polish Academy of Sciences Technical Sciences}, pp. e136751-e136751, 2021.
	
	\bibitem{RN72}
	G. Zan, D. J. Vine, R. I. Spink, W. Yun, Q. Wang, and G. Wang, "Design optimization of a periodic microstructured array anode for hard x-ray grating interferometry," \textit{Physics in Medicine \& Biology}, vol. 64, no. 14, p. 145011, 2019/07/16 2019, doi: 10.1088/1361-6560/ab26ce.
	
	\bibitem{RN73}
	G. Zan, D. J. Vine, W. Yun, S. J. Y. Lewis, Q. Wang, and G. Wang, "Quantitative analysis of a micro array anode structured target for hard x-ray grating interferometry," \textit{Physics in Medicine \& Biology}, vol. 65, no. 3, p. 035008, 2020/02/04 2020, doi: 10.1088/1361-6560/ab6578.
	
	\bibitem{RN74}
	F. Scholkmann, V. Revol, R. Kaufmann, H. Baronowski, and C. Kottler, "A new method for fusion, denoising and enhancement of x-ray images retrieved from Talbot–Lau grating interferometry," \textit{Physics in Medicine and Biology}, vol. 59, no. 6, pp. 1425-1440, 2014/02/28 2014, doi: 10.1088/0031-9155/59/6/1425.
	
	\bibitem{RN75}
	A. L. D. Cunha, J. Zhou, and M. N. Do, "The Nonsubsampled Contourlet Transform: Theory, Design, and Applications," \textit{IEEE Transactions on Image Processing}, vol. 15, no. 10, pp. 3089-3101, 2006, doi: 10.1109/TIP.2006.877507.
	
	\bibitem{RN76}
	T. Xiang, L. Yan, and R. Gao, "A fusion algorithm for infrared and visible images based on adaptive dual-channel unit-linking PCNN in NSCT domain," \textit{Infrared Physics \& Technology}, vol. 69, pp. 53-61, 2015/03/01 2015, doi: 10.1016/j.infrared.2015.01.002.
	
	\bibitem{RN77}
	H. Liu et al., "Multimodal Image Fusion for X-ray Grating Interferometry," \textit{Sensors}, vol. 23, no. 6, p. 3115, 2023, doi: 10.3390/s23063115.
	
	\bibitem{RN78}
	Z. Zuo, Y. Xiao, Z. Liu, B. Liu, and Y. Yan, "Discrimination of neutrons and gamma-rays in plastic scintillator based on falling-edge percentage slope method," \textit{Nuclear Instruments and Methods in Physics Research Section A: Accelerators, Spectrometers, Detectors and Associated Equipment}, vol. 1010, p. 165483, 2021/09/11 2021, doi: 10.1016/j.nima.2021.165483.
	
	
	
	\vspace{2cm} 
	\textbf{Haoran Liu} received a B.S. degree in nuclear engineering and nuclear technology from Engineering \& Technical College of Chengdu University of Technology, Sichuan, China in 2022. He is currently a doctoral candidate at Chengdu University of Technology. His research interests include computed tomography, 3rd-generation neural networks, deep learning, radiation detection, and digital image processing.
	

	\textbf{Mingzhe Liu }received a B.S. degree in computer application from Chengdu University of Technology, Sichuan, China, in 1994, and a M.S. and a Ph.D. degree in computer science from Massey University, New Zealand, in 2006 and 2010, respectively. He is currently a professor at the College of Computer Science and Cyber Security, Chengdu University of Technology. His research interests include intelligent information processing, information security, and system control.
	

	\textbf{Peng Li} is an undergraduate student in nuclear engineering and nuclear technology from Engineering \& Technical College of Chengdu University of Technology, Sichuan, China. His research interests include artificial neural networks and digital image processing.
	

	\textbf{Jiahui Wu} received a B.S. degree in nuclear engineering and nuclear technology from East China University of Technology, Jiangxi, China, in 2015. She's currently working on her master's degree at Chengdu University of Technology. Her research interests include medical imaging and processing techniques.
	

	\textbf{Xin Jiang} received a B.S. degree in information and computing science from the Chengdu University of Technology, Sichuan, China, in 2012, a M.S. degree in instrument and meter engineering, and a Ph.D. degree in nuclear technology and application from Chengdu University of Technology, Sichuan, China, in 2015 and 2020, respectively. His research interests include medical image processing and deep learning.
	

	\textbf{Zhuo Zuo} received a B.S. degree in nuclear technology from Chengdu University of Technology, Sichuan, China, in 2014, and a M.S. degree in nuclear technology and application from Chengdu University of Technology, Sichuan, China, in 2017. He is currently an associate professor at Engineering \& Technical College of Chengdu University of Technology. His research interests include nuclear detection and readout techniques.
	
	\textbf{Bingqi Liu }received a B.S. degree in automation from human university of arts and science, Hunan, China, in 2012, and a M.S. degree in instrumentation engineering and a Ph.D. degree in nuclear technology and applications from Chengdu University of Technology, Sichuan, China, in 2015 and 2019, respectively. He is currently an associate professor at the School of Mechanical Engineering, Chengdu University, and a postdoctoral researcher at the 209th Research Institute of China Ordnance Industry. His research interests include instrumentation and image processing.
	
	
\end{thebibliography}
\end{document}